%% file: ascexmpl-new.tex
\documentclass[Journal,letterpaper]{ascelike-new}
\pdfoutput=1
\usepackage[utf8]{inputenc}
\usepackage[T1]{fontenc}
\usepackage{lmodern}
\usepackage{graphicx}
\usepackage[figurename=Fig.,labelfont=bf,labelsep=period]{caption}
\usepackage{subcaption}
\usepackage{amsmath}
\usepackage{multirow}
\usepackage{newtxtext,newtxmath}
\usepackage[colorlinks=true,citecolor=red,linkcolor=black]{hyperref}
\usepackage{algorithm}
\usepackage[noend]{algpseudocode}
\usepackage{pdflscape}
\NameTag{Agapaki, \today}
\begin{document}
\nolinenumbers

\title{INSTANCE SEGMENTATION OF INDUSTRIAL POINT CLOUD DATA}

\author[1]{Eva Agapaki}
\author[2]{Ioannis Brilakis}

\affil[1]{Senior Software Developer, PTC Inc.,U.S.A. Email: agapakieva@gmail.com}
\affil[2]{Laing O’Rourke Reader, Department of Engineering, University of Cambridge, CB2 1PZ, U.K.}

\maketitle

\input{abstract}

\input{introduction}

\input{background}

\input{proposedSolution}

\input{results}

\input{conclusions}

\subsection{DATA AVAILABILITY}

Some or all data, models, or code used during the study were provided by a third party. Direct requests for these materials may be made to the provider as indicated in the Acknowledgements.

\subsection{ACKNOWLEDGEMENTS}

We thank our colleague Graham Miatt, who has provided insight, expertise and data that greatly assisted this research. We also express our gratitude to Bob Flint from BP International Centre for Business and Technology (ICBT), who provided data for evaluation. The research leading to these results has received funding from the Engineering and Physical Sciences Research Council (EPSRC) and the US National Academy of Engineering (NAE). AVEVA Group Plc. and BP International Centre for Business and Technology (ICBT) partially sponsor this research under grant agreements RG83104 and RG90532 respectively. We gratefully acknowledge the collaboration of all academic and industrial project partners. Any opinions, findings and conclusions or recommendations expressed in this material are those of the authors and do not necessarily reflect the views of the institutes mentioned above.

\input{figures}

\input{Tables}


\bibliography{references}

%
%
%

\end{document}

%% file: abstract.tex
\begin{abstract}
The challenge that this paper addresses is how to efficiently minimize the cost and manual labour for automatically generating object oriented geometric Digital Twins (gDTs) of industrial facilities, so that the benefits provide even more value compared to the initial investment to generate these models. Our previous work achieved the current state-of-the-art class segmentation performance (75\% average accuracy per point and average AUC 90\% in the CLOI dataset classes) as presented in \cite{agapaki2020cloi} and directly produces labelled point clusters of the most important to model objects (CLOI classes) from laser scanned industrial data. CLOI stands for C-shapes, L-shapes, O-shapes, I-shapes and their combinations. However, the problem of automated segmentation of individual instances that can then be used to fit geometric shapes remains unsolved. We argue that the use of instance segmentation algorithms has the theoretical potential to provide the output needed for the generation of gDTs. We solve instance segmentation in this paper through (a) using a CLOI-Instance graph connectivity algorithm that segments the point clusters of an object class into instances and (b) boundary segmentation of points that improves step (a). Our method was tested on the CLOI benchmark dataset \cite{Agapaki2019CLOI:Facilities} and segmented instances with 76.25\% average precision and 70\% average recall per point among all classes. This proved that it is the first to automatically segment industrial point cloud shapes with no prior knowledge other than the class point label and is the bedrock for efficient gDT generation in cluttered industrial point clouds. 
\end{abstract}

%% file: introduction.tex
\section{Introduction}
Maintenance and safety management are vital operations in the life-cycle of existing industrial facilities. Poor maintenance and safety deficiencies lead to equipment failure, which can have significant environmental, economic and societal impacts. In terms of economic impacts, unplanned downtime costs in the US due to corrective or poor maintenance are estimated to be \$50 billion per year with 44\% of all unscheduled equipment downtimes resulting from aging equipment \cite{NationalInstituteofStandardsandTechnology2018TheManufacturing}. Inefficient and ineffective facility management as well as poor documentation of the existing condition of facilities have been the primary causes of these issues. As a result, maintenance strategies take effect well after the problems have been identified. A quicker and more efficient maintenance scheme is a necessity generated by these market demands. 

The key for the prevention of the above-mentioned issues is proper refurbishment plans and preventive maintenance of industrial assets as addressed in recent studies. For instance, the Chartered Institute of Building in the U.K. estimates that refurbishing and retrofitting 93\% of existing industrial facilities will be a major focus by 2050 \cite{Edwards2011BuildingsRetrofit}. Another example of the perceived value of preventive maintenance proposed by the Association of Swedish Engineering Industries \cite{Bokrantz2016HandlingIndustry} is the strategy to eliminate production shutdowns in Sweden by 2030. Similarly, the German government has established the need of retrofitting existing buildings primarily to decrease their energy demands by 2050 \cite{Germanretrofit2011}. Analogous provisions will be established in the US, since over 70\% of existing retail buildings is built before 1980 \cite{USEnergyAdmin2006}. We argue that these market demands set the ground for the generation and maintenance of up-to-date DTs. However, gDTs are not currently used in most facilities. This occurs because the perceived cost of generating and maintaining the DT greatly surpasses the perceived benefits of the DT. The main reason for that is partly due to the high ratio of manual labor cost while generating the DT to data collection (laser scanning), which is roughly ten \cite{Lu2017RecursiveModelling,Hullo2015Multi-SensorArchitectures,Fumarola2011GeneratingApproaches,Agapaki2017PrioritisingTime,brilakis2019infrastructure}. This explains why there is an urgent need to develop less labor-intensive industrial modeling tools that can increase the productivity of industrial assets and their maintenance. In this paper, we address a core step of the digital twinning process, i.e instance segmentation. 

Leading 3D CAD vendors (Autodesk, AVEVA, Bentley, FARO and ClearEdge3D) have developed software that has a variety of 3D modeling features, which enable modeling from point cloud data, however none of these software assigned instance labels directly to individual point clusters. There needs to be an intermediate step first by fitting geometric shapes of which instance labels can be inferred.This process has two disadvantages:

\begin{enumerate}
  \item[\bf Drawback 1:] it inherently assigns some erroneous labels to points that do not belong to the cylinder instance. This is due to the reason that primitive shapes are perfect shapes, whereas physical objects are imperfect. Therefore, many more points that do not belong to the cylinder instance are assigned the wrong label due to fitting errors. On the other hand, class and instance segmentation give representation power to the TLS point clouds as suggested by \cite{Wang2019AssociativelyClouds}. This is because it embeds semantic information (class and instance labels) directly on the real point cloud representation, without the need to introduce shape primitives into the representation, as is required when using fitting methods or commercial software. In other words, segmentation methods bypass the stage of surface generation and directly output labelled point clusters. 
  \item[\bf Drawback 2:] the robustness of primitive fitting RANSAC-based methods is highly dependent on the spatial distribution of samples \cite{Liang2018DeepDetection}. In other words, samples with points that are closely located to each other usually cannot be properly detected \cite{Liang2018DeepDetection,Li2019PrimitiveSegmentation}. Many points that belong to separate instances may be grouped together in the same shape, especially if they are too close to each other.
\end{enumerate}

We showed in our previous work \cite{agapaki2020cloi,Agapaki2018PrioritizingFacilities,Agapaki2018State-of-practiceFacilities} that geometric modeling currently consists of three main steps: (a) primitive shape detection, (b) semantic classification of detected shapes and (c) fitting. We evaluated in \cite{Agapaki2018PrioritizingFacilities} state-of-the-art, semi-automated commercial packages and demonstrated that EdgeWise \cite{ClearEdge2019PlantCapabilities} automatically detects cylinders with 62\% precision and 75.6\% recall on average. This can be translated into 64\% of man-hour savings for cylinder modeling. However, taking an example of a small petrochemical plant with 240,687 objects and 53,834 pipes, 2,382 manual labor hours are still needed to model these cylinders \cite{Agapaki2018PrioritizingFacilities}. In summary, we have shown in \cite{agapaki2020cloi,agapaki2020scene} that the state-of-the-art 3D modeling practice has three main limitations: (a) the modelers fit standardized structural steel shapes after segmenting the structural elements manually or roughly selecting regions of interest using clipping polygons, (b) the modelers define parameters which determine cylinder detection and (c) EdgeWise does not directly assign class or instance labels per point. Rather it fits 3D solid standardized shapes and from them instance labels can be inferred. This process is error-prone especially in cases where two instances are very close or incorrect labels are assigned to instances due to fitting errors. It is easily distinguishable that the current practice still needs substantial manual efforts and is not designed to counteract the high costs of gDT generation. Yet another challenge in the adoption of the current state-of-the-art deep learning algorithms in the industrial sector for the generation of gDTs is the incompatibility of current Building Information Model (BIM) formats with the inputs these algorithms need \cite{SACKS2020100011}. The formats of BIM models are (a) proprietary file formats specific to BIM authoring platforms or (b) open IFC file formats on EXPRESS \cite{ISO2004}. The inputs that machine learning algorithms need to process are not compatible with either the complete models, partial assemblies of models or model components. Users would need to extract the relevant information anew for each use. Extraction of relevant objects and their properties requires parsing the IFC files or file formats. Translation of formats results in loss of information, which is a core problem of these gDT generation methods. This necessitates the requirement to redesign the procedure of gDT generation. 

We argue that cost reduction of gDT generation will be achieved by automating the following steps: (a) class segmentation, (b) instance segmentation and (c) fitting. We addressed class segmentation with our CLOI-NET framework estimating 70\% of labor hour savings in that task \cite{agapaki2020cloi}. Point cloud fitting to other geometric representations is a solved problem as proved by \cite{Agapaki2020}. However, instance level segmentation on industrial data remains an open problem and its value is of paramount importance since it can provide sufficient geometric information to generate the gDTs by fitting 3D objects to the instance point clusters. Class and instance segmentation definitions are provided below. 

Class segmentation describes the procedure of partitioning a laser scanned factory with class labels assigned per point (such as cylinder, elbow, I-beam, valve) \cite{Li2019ASegmentation}. Class segmentation has been widely studied in recent years \cite{arshad2019dprnet,che2019object,czerniawski2020automated,liang2019hierarchical,lu2020pointngcnn,Ma20183DRelationships,rethage2018fully,xie2019review,ye20183d,perez2017semantic,Dimitrov2016Non-UniformModeling,zhang2019review}. Since the scope of this paper is instance segmentation and not class segmentation, we choose to omit this discussion. The reader can refer to \cite{agapaki2020cloi} for a comprehensive literature review and also the state-of-the-art method for segmenting industrial shapes. Instance segmentation is closely related to class segmentation. While class segmentation assigns the same label to points that belong to different instances of the same class, instance segmentation assigns a label per point based on the individual object that the point belongs to. Instance segmentation is different from object detection, which is the procedure of identifying the location of objects that belong to a certain class without assigning a class label per point. Detected individual objects are usually represented by a bounding box that contains the object. 

This paper is the first to automatically generate instance point clusters from Terrestrial Laser Scanned (TLS) industrial data. We present our novel automated methodology in two parts: Step 1 predicts an instance label per point by using a graph-based method, namely Breadth First Search (BFS) that was originally introduced by \cite{Bauer1972TheLanguages}. Step 2 is a boundary segmentation method that is used to enhance the instance segmentation results of Step 1. We use the most labor intensive industrial object shapes to segment based on our previous work \cite{Agapaki2018PrioritizingFacilities}. These are in descending order of labor intensiveness: electrical conduit, straight pipes, circular hollow sections (CHSs), elbows, channels, solid bars, I-beams, angles, flanges and valves. Labor intensiveness is measured based on two indicators: frequency of appearance of the objects in as-designed BIM models and labor hours needed to manually model those.

We first elaborate on the state-of-research instance segmentation methods in the background section, which is followed by our proposed methodology. We then evaluate our methodology on a new point cloud dataset named ``CLOI'' that was introduced in our previous work \cite{agapaki2020cloi}. The results, achievements and limitations of our method are extensively discussed, and the final section follows with the conclusions, contributions and future directions for research.

%% file: background.tex
\section{BACKGROUND}
\label{background}

Class segmentation methods applied on industrial shapes have been widely investigated. \cite{Dimitrov2015SegmentationSystems,Perez-Perez2016SemanticSegmentation} use a region growing method and local features to refine the class labels of indoor point cloud scenes. A comprehensive review of class segmentation methods based on hand-crafted features is provided by \cite{agapaki2020cloi}.

\subsection{Instance Segmentation}

The sections that follow analyze the metrics and state-of-the-art methods currently used to segment instance point clusters from TLS data.

An object proposal is marked as detected (true positive) if the Intersection over Union (IoU) is above a certain threshold. The predicted labels are then compared with the ground truth labels pointwise and precision, recall and Intersection-over-Union ($IoU$) scores are computed. The IoU score is defined as:

\begin{equation}
    IoU_c(pred_{ins_i},gt_{ins_j})\geq t
    \label{IoU_ins}
\end{equation}

where $pred_{ins_i}$ and $gt_{ins_j}$ correspond to the predicted and ground truth point clusters respectively, for $1\leq i\leq N$ and $1 \leq j \leq M$.

$pred_{ins_j}$ and $gt_{ins_j}$ match if and only if Equation~\ref{IoU_ins} holds for some predetermined threshold $t$.

There are three properties that the $pred_{ins_i}$ and $gt_{ins_j}$ need to comply with:

\begin{enumerate}\setcounter{enumi}{0}
  \item For any $i,j$, $pred_{ins_i}\subseteq S$ and $gt_{ins_i}\subseteq S$, where S is the set of input points. Also, for any $i\neq i'$ and $j\neq j'$ we have $pred_{ins_i}\cap pred_{ins_{i'}} = \emptyset$ and $gt_{ins_j}\cap gt_{ins_{j'}} = \emptyset$. In other words, $gt_{ins_j}$ and $gt_{ins_{i'}}$ are disjoint sets of points. The same property applies for $pred_{ins_i}$ and $pred_{ins_{i'}}$.
  \item $\bigcup_{j=1}^M gt_{ins_j} = S$ meaning that every point of the input TLS industrial dataset belongs to some instance. This means that every point is assigned to an instance and those that do not belong to the \textit{CLOI} class instances are assigned to the ``other/clutter`` class.
\end{enumerate}

Instance recall and precision are measured with the same equations as defined in \cite{agapaki2020cloi}. The former measures the percentage of $gt_{ins_j}$ that is matched with some $pred_{ins_i}$, whereas the latter measures the percentage of $pred_{ins_i}$ that matches some $gt_{ins_i}$.

It is important to note that we follow the segmentation gDT generation strategy as it was proved to be more efficient in our previous work \cite{agapaki2020cloi} compared to object detection gDT strategy. Instance segmentation is the last step of this strategy before fitting geometric shapes to the point clusters. Deep learning is the core technique for instance segmentation tasks. Instance segmentation methods are a relatively new research field in the computer vision community that is greatly developing in the last years adopting the advances in instance segmentation of images. Instance segmentation has been widely used in image processing \cite{He2017MaskR-CNN,Newell2016StackedEstimation,Girshick2014RichSegmentation,Newell2017AssociativeGrouping}. 
The readers can refer to \cite{Huang2017Speed/accuracyDetectors} for a comprehensive comparison between Faster R-CNN and other frameworks. 3D instance segmentation methods in research related areas are then investigated, since there is no method that has segmented instances on industrial facility scenes. 

\subsection{3D Instance segmentation methods in research related areas} 

3D instance segmentation has attracted a lot of research attention recently \cite{Wang2018SGPN:Segmentation,Wang2019AssociativelyClouds,Yi2019GSPN:Cloud,Pham2019JSIS3D:Fields}. Researchers segment 3D instances by learning to group per-point features \footnote{A feature vector is assigned per point.} into instances. These networks usually use PointNET/PointNET++ to project the 3D points in a high-dimensional feature vector \cite{Qi2017PointNET:Segmentation,Qi2017PointNet++:Space}. We group the instance segmentation methods into two classes: (a) {\bf shape based} (top-down) and (b) {\bf shape-free} (bottom-up) using a similar grouping with \cite{Wang2019AssociativelyClouds}. 

Shape-based methods generate object proposals (such as 3D bounding boxes) in a similar way like 2D object proposals on images \cite{He2017MaskR-CNN}. The generation of bounding boxes is an intermediate step in order to extract instance labels per point. Shape-free methods learn to associate pixel-predictions or 3D point predictions to object instances. Region-based CNNs (R-CNNs) \cite{Girshick2014RichSegmentation} are the baseline of instance segmentation methods. They have two main steps: (1) prediction of a set of candidate object proposals (mask proposals) and (2) training of an object classifier to choose the best candidate proposal (final instance masks). Shape-based methods work robustly with images, however these techniques are challenging on TLS point clouds. The main challenge shape-based methods have on 3D data is that they are not trained from scratch on 3D object detectors rather they make use of image features \cite{Deng2017AmodalImages,Liang2018DeepDetection,Qi2018FrustumData}. Recently, 3D instance segmentation of multi-modal inputs (RGB-D scans) was studied by \cite{Hou20193D-SIS:Scans} achieving a combination of 2D and 3D feature learning. A recent network (3D-BoNet) directly learns 3D bounding boxes for all instances in a TLS point cloud dataset by using two parallel network branches for a) instance-level bounding box and 2) point mask prediction \cite{Yang2019LearningClouds}. Their method is designed to learn object boundaries. The main drawback of 3D-BoNET is that drawing 3D bounding boxes for instances of industrial objects is a challenging task given the high variability in the dimensions of \textit{CLOI} shapes. Another challenge is that there is no direct link between the predicted boxes and the ground truth labels to supervise the network, even though the 3D-BoNET enforces a set of geometric rules to predict the 3D bounding boxes.

Another example of shape-based methods is GSPN that stands for Generative Shape Proposal Network \cite{Yi2019GSPN:Cloud}. GSPN generates object proposals by understanding the underlying object geometry. For example, GSPN does not produce blind proposals that contain multiple objects or parts of an object. It takes a point cloud and a seed point as input, uses a Conditional Variational Autoencoder (CVAE) to generate a proposal, converts the proposal to a ROI box and then uses an R-PointNET++ (mask R-CNN) to segment an object. Only the candidate bounding boxes that have confidence score greater than 50\% are used, similarly to other instance segmentation methods. The main drawback of this method is that industrial shapes have different geometric shapes and are more intricate than everyday objects, as such the viability of this method on industrial data is limited.

On the other hand, shape-free approaches generate instances based on feature aggregation. SGPN (Similarity Group Proposal Network) trains a network that outputs instance level labels per point, based on a similarity matrix between all pairs of points. This matrix compares the feature vectors of 3D points and assigns them to instances. SGPN tests three potential similarity classes for each pair of points: (a) points belong to the same object instance, (b) points share the same class label but do not belong to the same object instance and (c) points do not share the same class label. ASIS (Associatively Segmenting Instances and Semantics) is another network that associates semantic and instance level predictions \cite{Wang2019AssociativelyClouds}. The ASIS network is trained to segment 3D points into instances by comparing the feature vectors of those points projected in an embedding space. The points that are close to each other in the embedding space belong to the same instance, whereas points belonging to different instances are apart. 

JSIS3D is another shape-free instance network that couples the class and instance segmentation tasks \cite{Pham2019JSIS3D:Fields}. They achieve that in two steps. First, they apply a multi-task network based on PointNET \cite{Qi2017PointNET:Segmentation} that predicts the class and instance labels per point and secondly they optimise those labels by developing a multi-value CRF model. The 3D-BEVIS (3D bird's-eye-view instance segmentation) predicts an instance and class label given a TLS point cloud applying three stages \cite{Elich20193D-BEVIS:Segmentation}: 1) a 2D instance feature network that learns instance features from a bird's eye view (top view of a scene), 2) after concatenating the instance features to the TLS point cloud features, a 3D feature propagation network propagates and predicts instance features for all points in the scene and 3) final prediction and clustering of the class and instance labels. The main limitation of this network is that it is not suitable for vertically oriented objects since they are not visible in the 2D bird's eye view representation. Overall, the ASIS \cite{Wang2019AssociativelyClouds}, JSIS3D \cite{Pham2019JSIS3D:Fields} and 3D-BEVIS \cite{Elich20193D-BEVIS:Segmentation} networks apply the same per-point feature grouping pipeline as initially proposed by SGPN \cite{Wang2018SGPN:Segmentation} to segment 3D instances. 

Shape-free instance segmentation methods are suitable for addressing in part our research problem, since 3D bounding box training on complex industrial scenes is far from a feasible task. Also, the high level of occlusion and density of objects in industrial scenes prevents the use of methods that rely on 3D projections in the 2D space like the shape based methods. However, shape-free instance segmentation methods also have disadvantages. The disadvantage of those methods is that they inevitably require a post-processing step such as mean-shift clustering \cite{Comaniciu2002MeanAnalysis} to segment the final instance labels which is a computationally intensive task. Another disadvantage of these methods is that they do not explicitly detect the object boundaries of objects, which is of significant importance for industrial shapes that have complex boundaries. 

The above-mentioned networks have only been used on point clouds of residential or office spaces \cite{Armeni20163DSpaces} and urban scenes \cite{Behley2012PerformanceEnvironments,Hackel2017SEMANTIC3D.NET:Benchmark,Munoz2009ContextualNetworks,Roynard2018Paris-Lille-3D:Classification,Steder2010RobustFeatures,Geiger2012AreSuite}. Instance segmentation of industrial point clouds has not been achieved so far. Traditional computer vision methods, such as region growing and hand crafted features have been investigated to segment Mechanical, Electrical and Plumbing (MEP) systems in TLS point clouds \cite{Dimitrov2015SegmentationSystems}. This method takes point cloud density, surface roughness, curvature and clutter into consideration. Although the main limitations are (a) over segmentation especially for highly occluded scenes and (b) lack of contextual inter-connectivity relationships to connect shapes, principal curvature is a local feature that can describe the 3D structure of points in occluded scenes and is useful for segmenting point cloud scenes. We investigated that feature in the CLOI-NET class segmentation method \cite{agapaki2020cloi}. The difference of traditional computer vision methods compared to our approach is that it is not dependent on class-specific geometric features and thus might be more applicable to new or unseen industrial classes. As mentioned above, shape-free instance networks are more appropriate for instance-level predictions of industrial environments due to the complex boundaries between industrial shapes that cannot be approximated by shape-specific object proposals (e.g. 3D bounding boxes).  

\subsection{Gaps in knowledge, objectives and research questions}

Considering the state of practice and body of research reviewed above, existing approaches attempt to generate instance point clusters using shape-based and shape-free approaches. These approaches have only been applied for indoor and urban scenes. As reviewed in \cite{agapaki2020cloi}, semi-automated software packages like EdgeWise (i) have only automated the detection of cylinders achieving 75\% recall and 62\% precision on average, (ii) manually detect the rest of the most important industrial shapes and (iii) not further classify cylinders into conduit or CHSs or pipes following the segmentation strategy. Many researchers have addressed instance segmentation of indoor buildings or offices, however there is no method available for industrial data. The sub problems that stem from these observations give a clear understanding on the exact pain points (Gaps in Knowledge: {\bf GiK}) and the research questions that need to be addressed in order to solve instance segmentation of these shapes.

A problem that was identified is that existing methods fit idealized industrial shapes that incur an extra error in the model (human-related or algorithmic-related detection error). The industrial shapes will never be in their exact location, but they will be in an approximate location. The only way to identify the exact locations of 3D shapes and directly capture the existing conditions is on the TLS dataset itself. {\bf GiK}: Current methods rely on other intermediate 3D representations (e.g. fitting standardized shapes in the point cloud) instead of directly processing the input point cloud data and segmenting it to individual instances.   

\begin{enumerate}
  \item[\bf RQ1:] How to automatically segment instances from the TLS industrial dataset without directly enforcing other 3D geometric representations? 
  \item[\bf RQ2:] How to perform instance segmentation at optimal performance compared to state-of-the-art instance segmentation methods? 
\end{enumerate}

%% file: proposedSolution.tex
\section{Proposed solution}
\label{proposed}

In this section, we derive our proposed solution by first defining our assumptions, then evaluating the state-of-research instance segmentation methods based on the literature review, scope and assumptions of our segmentation application on industrial scenes. We then present each step of our method and derive our hypothesis that will be used to validate the feasibility our method in the experiments section.

\subsection{Assumptions}

We assume the proposed CLOI-Instance segmentation method is feasible in the context of gDT generation under the following conditions:

\begin{enumerate}\setcounter{enumi}{0}
  \item [{\bf A1.}] The input data in the proposed method is ideal class segmentation labels, so that abnormal noisy points and outliers have been manually removed.
  \item [{\bf A2.}] The points at the interfaces between the instances are considered boundary points. This is expected to assist the segmentation of instances by clearly defining boundaries between instances.
  \item [{\bf A3.}] The class labels assist the instance predictions of the CLOI-Instance method. However, the instance segmentation task should be independent from the class segmentation method. For example, it is expected that it is a much easier task to assign the instance label to a point cluster (e.g. cylinder1) whose class label is known (e.g. cylinder). In other words, knowing the class label of an instance a-priori can assist the instance segmentation task, but the instance segmentation results should not impact or alter the class segmentation processing. Rather the class labels can benefit the instance segmentation, but not vice versa.
  \item [{\bf A4.}] The \textit{CLOI} instance segmentation is not only dependent on the geometric shape but also on the distance between connected points. For example, closely located shapes are expected to be more difficult to segment.
\end{enumerate}

Our proposed solution is based on these assumptions. The {\bf objective} of this paper is to provide an automated method to segment instances (individual shape point clusters) from the class labelled point clusters and to let the instance segmentation benefit from the class segmentation method by receiving it as an input. This objective is achieved by answering the {\bf research questions} {\bf RQ1} and {\bf RQ2}. 

The {\bf inputs} of our method are the ground truth class segmented point clusters. The ground truth is used, in order to evaluate the method on its own without adding the error of the class segmented predictions of our previous work \cite{agapaki2020cloi}. The combined problem of computing both class and instance labels is not within the scope of this paper. The outputs of our method are point-wise instance labels (individual point clusters of \textit{CLOI} shapes).

We aim to (1) extract the geometric features of the seven \textit{CLOI} types of labelled point clusters constituting an industrial point cloud and (2) use these features to segment into individual point clusters. The seven \textit{CLOI} categories are cylinders, elbows, channels, I-beams, angles, flanges and valves. The geometric features that are relevant to instance segmentation are the minimum number of points per instance and the minimum distance between instances (optimal trade-off between distance of neighboring instances and point density). To this end, we propose a novel instance segmentation method, which is a variation of region growing methods and uses geometric features (the minimum number of points per instance and minimum distance between instances) and class labels of points. The class segmentation labels refer to the \textit{CLOI} class labels as defined by \cite{agapaki2020cloi}.

Our method needs to automatically and rapidly generate individual point clusters where Oriented Bounding Boxes (OBBs) or Swept Solid shapes can be fitted should the users still prefer to use an ideal gDT for visualisation purposes. The reason for segmenting instances from the original point cloud data is to accurately capture the existing shapes rather than directly fit standardized solid shapes in IFC \cite{BuildingSMART2018InternationalOpenBIM} or related format. These solid shapes are so generic that even extruded or free-form solids representing specific template profiles (i.e. I-beams) are not representative but an estimate of the existing conditions. This is particularly important for pipelines since they are often corroded or rusty. 3D patches should be fitted to the extracted instance point clusters in order to have representative 3D geometry. This part is out of the scope of this paper. Therefore, we contend that instance segmentation to accurately generate instance point clusters is an essential step prior to fitting.

More specifically, the key {\bf hypotheses} of the instance segmentation method that need to be proved are presented below. 
\begin{enumerate}\setcounter{enumi}{0}
    \item The CLOI-Instance segmentation method is not significantly biased in the segmentation performance (mPrec, mRec, mIoU) for different facilities.
    \item The CLOI-Instance segmentation method has better performance in mPrec, mRec and mIoU compared to the current state-of-the-art practice.
\end{enumerate}

These hypotheses will be tested in the experiments and results section with the \textit{CLOI} benchmark dataset.

\subsection{Overview} 

~Figure \ref{fig:insmethodology} illustrates the workflow of the proposed methodology. The method uses the seven most important \textit{CLOI} classes as identified in \cite{Agapaki2018PrioritizingFacilities}. The inputs of the method are the spatial coordinates of the point clusters. The same 3D block generation method from \cite{agapaki2020cloi} is used for segmenting the input data. In other words, the initial TLS industrial data is partitioned in 3D non-overlapping sliding windows with overlapping 3D blocks. The method consists of two major steps: Step 1 predicts an instance label per point by using a graph-based method, namely Breadth First Search (BFS) that was originally introduced by \cite{Bauer1972TheLanguages}. Step 2 is a boundary segmentation method that is used to enhance the instance segmentation results of Step 1. 

The selection of our method is based on experimental evaluation of the state-of-the-art instance segmentation deep learning networks that are suitable for the application in industrial scenes and is presented in the paragraphs that follow. The suitability of the method is also assessed based on the following criteria that make industrial environments challenging scenes for instance segmentation tasks. These are:

\begin{enumerate}\setcounter{enumi}{0}
  \item 3D object boundaries are complex and cannot be approximated with bounding box lines. They are usually curved surfaces (i.e. elbows) or multi-faceted surfaces (i.e. valves).
  \item Industrial objects are very close in distance ($mm$ scale) and even overlap each other (e.g. angles of a steel bracing frame).
\end{enumerate}

The only methods that comply with the above-mentioned criteria are the shape-free instance segmentation methods as discussed in the literature review and the best performing networks on the Standford 3D indoor dataset \cite{Armeni20163DSpaces} are the SGPN \cite{Wang2018SGPN:Segmentation} and ASIS \cite{Wang2019AssociativelyClouds} networks. Therefore, we first investigated those. The performance of these networks on the oil refinery dataset (which is part of the ``CLOI'' dataset \cite{agapaki2020cloi} and will be introduced in the next section) is summarized in ~Table \ref{table:ASISSGPN}. The metrics used to evaluate the state-of-the-art networks are precision (Prec), recall (Rec) per \textit{CLOI} shape, mean precision (mPrec) and mean recall (mRec). It is important to note herein that an instance prediction is often considered correct if 50\% or more of the union of the points of the predicted instance and the ground truth instance overlap, following recent literature \cite{Hariharan2014SimultaneousSegmentation}. In other words, the IoU score between the predicted and the ground truth instance is at least 50\%. The same threshold is used for the evaluation of the SGPN and the ASIS networks. 

The results illustrated in ~Table \ref{table:ASISSGPN} show that SGPN has very low performance on the oil refinery data and ASIS performs better in all efficiency metrics. The results and especially Rec results show that the performance of these networks is relatively low for the gDT generation process with most \textit{CLOI} shapes being incorrectly segmented as ``other''. However, the ASIS network seems to have more promising performance, as such it was chosen as the baseline for further experiments that are explained in Step 2 of the instance segmentation method.

The results of the state-of-the-art instance segmentation deep learning networks demonstrate the following two pain points:

\begin{enumerate}\setcounter{enumi}{0}
  \item Instance segmentation is a more complex task compared to class segmentation. Class segmentation only accounts for the object type that each point belongs to (e.g. cylinder point). Instance segmentation not only assigns the label of the class that each object belongs to but also it separates the individual instances from each other (e.g. cylinder 1). This can be very challenging especially for instances that are too close to each other, have complex boundaries or are cluttered. Therefore, it should be ``informed'' from the class labels and should not be independent ({\bf P1}).
  \item The boundaries of instance point clusters are not explicitly enforced in the existing instance segmentation networks. As a result, instances are not properly segmented as demonstrated in ~Table \ref{table:ASISSGPN} ({\bf P2}).
\end{enumerate}

Taking into consideration the factors discussed above, we test a deep learning network in Step 2 of the method that classifies every point as boundary and non-boundary, named Boundary-NET. The architecture of this network is based on the geometric deep learning network ASIS \cite{Wang2019AssociativelyClouds} and PointNET++ \cite{Qi2017PointNet++:Space}.

Another factor to consider for robust instance segmentation in industrial facilities is the shape scalability. Industrial shapes and specifically the \textit{CLOI} shapes that are of interest in this paper have different scales ranging from a few centimeters (steelwork) to some meters (cylinders). More details about the \textit{CLOI} shape scalability can be found in \cite{agapaki2020cloi}. This illustrates the need for a method that is scale invariant. 

Scalability is a limitation of instance segmentation networks. Since the above-mentioned networks capture local geometric features of points at $1m^3$ 3D blocks, instances larger than that cannot be segmented. In other words, these deep learning networks do not scale efficiently with the number of points, that is why the ``block'' technique is applied. On the other hand, graph-based methods, such as Breadth First Search (BFS) are scalable, which means that these algorithms can process the entire TLS dataset without the need to subdivide it into 3D ``blocks''. Henceforth, the proposed CLOI-Instance method incorporates a graph-based method that can segment larger scale (more than $1m^3$ in volume) shapes.

The sections that follow describe each step of the CLOI-Instance method in detail.

\subsection{Step 1 - Graph-based connectivity of industrial shapes} 
\label{BFS}

We first use a Breadth First Search (BFS) algorithm to cluster points into instances that address the two pain points presented above. The method takes as input the boundary labels per point from the Boundary-NET network and the subsampled point cloud according to the farthest point sampling method described in \cite{agapaki2020cloi} for the 3D block generation. The method then assigns the same instance to two points that have the same instance label if their distance is less than a threshold value $t$ and these points belong to the same object type. These points form a graph of connected components. The algorithm outputs each individual connected component as an instance. The algorithm is presented in Algorithm~\ref{BFS_algo}. It constructs a graph $G(V,E)$, where $V$ are the vertices and $E$ is the set of edges that connect the vertices of the graph. Here we define $V = \{ 0, \dots N-1\}$ and $E = \{ (i,j) \ | \ 0 \leq i,j \leq N-1 \ and \ d_{i,j} < t \} $, where $d_{i,j}$ is the distance between the two points $i$ and $j$, $t$ is a threshold distance to split the instances and $N$ is the number of subsampled points of a 3D window. This means that two points $i$ and $j$ belong to the same instance if the following conditions are met: 

\begin{enumerate}\setcounter{enumi}{0}
    \item there is a path connecting $i$ and $j$ in the graph $G$ without containing any boundary points and such that all points in the path belong to the same class.
    \item $i$ and $j$ do not belong to a boundary, but are interior points. This condition will be explained in Step 2 of the method.
\end{enumerate}

Points belonging to a boundary are assigned to the closest instances. Algorithm~\ref{BFS_algo} is an algorithm with complexity $O(N*number\ of\ points\ in\ neighbourhood)$, given that neighbourhoods for each point have already been computed using a k-D tree structure for the nearest neighbour points \cite{Maneewongvatana2002AnalysisSets}. 

\begin{algorithm}
\caption{BFS Algorithm}\label{BFS_algo}
\hspace*{\algorithmicindent} {\textbf{Input:} $(X_i,Y_i, Z_i)=P_i$, $i \in [0, \dots ,N-1]$, $S_i = \{ {j \in \{0, \dots ,N-1\}\ |\ j \neq i} \}$, t} \\
 \hspace*{\algorithmicindent} {\textbf{Output:} array INS[i] with instance label of point $i$} 
\begin{algorithmic}[1]
\Procedure{Find connected point cluster components}{}
\State tree = kDtree(P)
\For {$i = 1 \dots N$}
\State L = query(tree,i,t)
\For {$j \ in \ L$}
\If {i $<$ j and $class_i$ = $class_j$ and $boundary_i = interior$ and $boundary_j = interior$}
\State INS[i] = join(i,j)
\EndIf
\EndFor
\EndFor
\State join each boundary point with the closest instance
\EndProcedure
\end{algorithmic}
\end{algorithm}

This algorithm encodes contextual information between industrial shapes in point cloud data. It should be noted that the distance function $d(i,j)$ is symmetric. This means that $d(P_i,P_j) = d(P_j,P_i)$, both indicating the distances of points $i$ and $j$. The distance $d$ between the points is the usual Euclidean distance.

We first evaluate the BFS method on its own and ~Table \ref{table:beforeafterresults1} demonstrates that although the BFS method performs better compared to the deep learning instance segmentation networks presented above, the performance can still be improved. 

Two parameters are fine-tuned to evaluate the performance of the BFS algorithm per \textit{CLOI} class. These are:

\begin{enumerate}
    \item[(a)] the minimum neighbourhood distance between two points ($\epsilon$) and
    \item[(b)] the minimum neighbourhood instance size ($\mu$).
\end{enumerate}

The minimum IoU thresholds that are used are 25\%, 50\% and 75\% as proposed by \cite{Wang2019AssociativelyClouds}. In other words, the performance of the BFS algorithm for these sets of parameters is measured at three levels of strictness: (a) IoU threshold of 25\% (b) 50\% and (c) 75\%. Inevitably, when the minimum IoU threshold decreases, the BFS algorithm has softer criteria to correctly segment the instance point clusters. Therefore, the instance segmentation performance is expected to be better. 

The selection of the minimum instance size is determined by experiments for minimum IoU threshold of 50\% as shown in ~Figure \ref{fig:minInsSize}. We test various values for the minimum instance size ranging from 10 points per instance to 200 points per instance. The trend in ~Figure \ref{fig:minInsSize} shows that the larger {$\mu$} is, the higher the precision and the lower the recall of the BFS algorithm. The same trend is observed in all \textit{CLOI} facilities. This is attributed to the fact that the more points instances are constrained to have, the higher the chance that those points will represent these instances accurately. As the minimum instance size ($\mu$) is increased, the number of true positives decreases since ground truth instances with few points might be ignored from consideration and at the same time precision will be increased because small instances have a larger probability of being noise. Therefore, there is a trade-off between precision and recall, depending on the minimum size of points ($\mu$) that an instance can have. The optimal minimum instance size is 20 points based on ~Figure \ref{fig:minInsSize}. 

Representative results to find the minimum neighbourhood distance between two points ($\epsilon$) for minimum instance size ($\mu$) of 20 points that optimize the performance per class are given in ~Table \ref{table:beforeafterresults1}. This table shows that the larger the radius, the lower the Rec of cylinders. This means that some cylinder instances are too close to each other and the BFS algorithm considers those as a single cylinder. This is particularly important for the case of conduits that are too close even in most cases overlapping each other. The selection of the radius $\epsilon$ is in accordance with the spacing specifications between pipelines \cite{Beale2010ProcessDesign}. Particularly, the minimum spacing between the center lines of pipelines with and without flanges is 2.5cm and 5cm respectively. The range of the spacings though varies based on the pressure of the fluid the pipeline carries. The spacing between the center lines of parallel conduits is 5cm \cite{Hensley2017DavsBook}. For other conduit configurations, the configuration of the conduits determines the spacing requirement between them (i.e. 90$^{\circ}$ bends, 45$^{\circ}$ bends with obstructions). ~Table \ref{table:beforeafterresults1} shows that the other \textit{CLOI} classes have higher Rec for $\epsilon \leq 4cm$ and then lower Rec after $\epsilon = 4cm$. This is evident since these classes are distant from each other. For example, it is rare to find a valve too close to another valve. The Prec metric increases for all \textit{CLOI} classes. 

The set of parameters with optimal precision and recall performance are $\epsilon=0.04m$ and $\mu=20$ points based on the observations above. Increased recall is more important than precision when segmenting instances, since the more instances are correctly predicted the less manual labour hours are needed to identify the missing instances. On the other hand, the user can discard more quickly the wrongly predicted instances rather than manually segmenting instances that were not predicted. 

The performance of the BFS algorithm with the selected optimal parameters given the ground truth class labels is 65.2\% mPrec and 57.1\% mRec when testing on the oil refinery dataset. It is worth mentioning that the mean metrics are computed given the performance of the BFS algorithm in the seven \textit{CLOI} classes excluding the ``other'' class for IoU threshold of 50\%.

\subsection{Step 2 - Boundary segmentation} 

We then introduce a classification of points into \textit{boundary} and \textit{non boundary} points and then use a PointNET-based network architecture to perform binary classification in order to address the pain point {\bf P2} and improve the BFS performance from Step 1. A {\bf non-boundary point} is defined as a point whose neighbourhood has points that only belong to one instance. If a point of another instance is encountered in the neighbourhood, then the original point is a {\bf boundary}. These geometric features (boundaries) are then used in the BFS algorithm that was explained in Step 1. 

Intuitively, the proposed Boundary-NET network has similar architecture with the ASIS network \cite{Wang2019AssociativelyClouds}. However, Boundary-NET accounts for the instance segmentation separately from the class segmentation given the assumption that class labels are correctly assigned to each point. This modification aims to solve the pain point {\bf P1} that was identified by the experiments presented above based on the performance of these networks on the class segmentation point label predictions. The ASIS network and the Boundary-NET network architectures are illustrated in Figure~\ref{fig:ASIS_network}(a) and Figure~\ref{fig:ASIS_network}(b) respectively. In other words, the method detaches the class and instance segmentation branches of the ASIS network, so that the class segmentation branch boosts the performance of the instance segmentation branch, while the influence of the instance segmentation branch should not impact the class segmentation labels. We could have chosen to completely remove the class label learning from the network, but observed improved performance when the class label learning was kept. In other words, these networks give high instance segmentation performance if the class segmentation is perfect (Table \ref{table:ASISSGPN}). It is important to note that the use of predicted class labels as input in the instance segmentation method deteriorates performance significantly, which proves that the state-of-the-art instance segmentation networks are very sensitive to noisy and imperfect class labels.

An illustration of the full pipeline of the network is presented in Figure~\ref{fig:ASIS_network}(c). This network takes as input the spatial coordinates of 3D points of each block and samples points using a PointNET++ \cite{Qi2017PointNet++:Space} sampling layer to account for spatial features between the points. Analysis of the PointNET++ \cite{Qi2017PointNet++:Space} sampling layer is beyond the scope of this paper and the readers can refer to \cite{Agapaki2020} for further details. Then, another PointNET++ layer is used to propagate features of all points. The Boundary-NET module aggregates the features from the instance and class segmentation layers. The method obtains the labels per point after a training procedure that minimises the cross entropy loss $H_p(q)$ for this binary classification Boundary-NET network, which is given by:

\begin{equation}
    H_p(q) = -\sum_{i=1}^N y_i \log p(y_i)  + (1-y_i)\log(1-p(y_i))
\end{equation}

where $y$ is the label (1 for boundary point, 0 for interior points) and $p(y)$ is the probability of the point being boundary for each point.

The points are classified by using a softmax function in order to predict the boundary label. The predicted boundary labels are then used in Step 1 to predict the instance labels per point.

The Boundary-NET network as proposed above classifies each point as boundary or interior, however there are still instances located too close to each other (so boundaries are not clearly distinct) or noise impedes joining instances together or segmenting them as shown in Figure~\ref{fig:BFS_noiseproximity_gt}. This figure illustrates the ground truth and predicted instances after applying the Boundary-NET network on the petrochemical and the oil refinery TLS datasets. Instances are distinguished based on their colour. The results show that the performance of the Boundary-NET network is low. The primary reason for that is the unbalanced classes of boundary and non-boundary points. In other words, the boundary points are much less compared to the non-boundary points and this biases the performance of the network that cannot correctly predict the boundary points. Therefore, we then propose the use of \textit{class boundary} points. A point is defined in this work as \textit{class boundary} if and only if around the neighbourhood of that point there is at least one point with a different class label. The ground truth class labels will be used for the validation of the CLOI-Instance method. 

The boundary segmentation method is not sufficient on its own to perform accurate instance segmentation. The reason for that is that instances with the same class are closely located to each other as well. Therefore, the combination of the boundary segmentation method and the BFS algorithm from Step 1 is proposed in Figure~\ref{fig:insmethodology} to boost performance. 

We therefore propose a hybrid method that has two main steps: a graph-based segmentation algorithm, BFS and a boundary segmentation method. The former takes as input the entire industrial point cloud and generates connected components based on connectivity relationships in order to segment the instances as output. For the latter, a boundary segmentation method is proposed that takes as input the class-labelled points from the class segmentation step and outputs binary labels on whether a point is a boundary point or not. The novelty of the method is two-fold: 

\begin{enumerate}\setcounter{enumi}{0}
    \item the efficiency of the BFS algorithm by applying it on the entire point cloud and connectivity between points
    \item the intelligence of the boundary segmentation method to account for boundary points and robustly process points in small regions.
\end{enumerate}

%% file: results.tex
\section{EXPERIMENTS AND RESULTS}
\label{methodology}

\subsection{GEOMETRIC PARAMETER SEARCH}
\label{researchActivities}

The research design that we followed is to validate each process of the instance segmentation methodology and to find the optimal geometric parameters of each method in our validation dataset.We validated the instance segmentation method in the four real-world industrial facilities of the \textit{CLOI} dataset. These were originally introduced by \cite{agapaki2020cloi}. The number of instances and number of points per class are presented in Table \ref{table:CLOIstats}.

Experiments were first conducted on the BFS algorithm (Step 1) of the instance segmentation method to validate the suitability of the selected parameters on the \textit{CLOI} dataset (minimum instance size, $\mu=20$ and minimum neighbourhood distance between two points, $\epsilon$). The next step was to use the boundary segmentation method (Step 2) in order to enhance the BFS algorithm from Step 1. 

We revise the two limitations of the CLOI-Instance methodology with randomly selected parameters. These are summarised as follows:

\begin{enumerate}\setcounter{enumi}{0}
  \item The BFS algorithm oversegments sparsely scanned regions. For instance, in these cases, the object may be cut in more than one instances. 
  \item Instances that are located too close to each other will be predicted as the same instance. The reader can refer to ~Figure \ref{fig:BFS7cm} for an illustrative example of this case ({\bf BFS2}).
\end{enumerate}

Therefore, further experiments were conducted on the BFS algorithm to define the parameters that can mitigate the {\bf BFS1} limitation presented above on the \textit{CLOI} dataset. We found the lower bound value of the radius for which the BFS algorithm can segment one object at a time. In other words, each instance point cluster is considered individually as input and only runs the BFS algorithm on that instance. If the radius is too small, the BFS algorithm will split an instance in many sub-instances. If the radius is too large, the BFS algorithm may cover the instance of interest and other closely located instances. The performance of the BFS algorithm on the radius selection is measured with the $mRec_{ins}$ metric. In other words, this metric measures the number of instances segmented with IoU threshold greater than specific threshold values over the total number of ground truth instances. ~Figure \ref{fig:BFSradius} summarises the results for each \textit{CLOI} facility. The results show that if the radius of the neighbourhood ($\epsilon$) increases, then the performance ($mRec_{ins}$) increases. Considering the elbow method \cite{Ketchen1996TheCritique}, a good lower bound $\epsilon$ for all facilities is $\epsilon \geq 3cm$. The poor performance of a radius of around $1cm$ indicates that there are gaps between the 3D points of $1cm$ within the same instance, whereas gaps of $4cm$ are not that frequent. The criterion for the radius selection is having $mRec_{ins}$ $\geq$ 90\% at IoU=50\%. This consistently results in very high $mRec_{ins}$ for all the \textit{CLOI} facilities and all the IoU thresholds (25\%, 50\% and 75\%), which is close to or even above 80\%. 

Then, we investigated the suitability of the lower boundary $4cm$ neighbourhood radius ($\epsilon$) per \textit{CLOI} class and each \textit{CLOI} facility for every IoU threshold in Figure~\ref{fig:BFSradiusshape}. The results show that the same trend is observed for each class. This work introduces the $Rec_{ins}$ metric, which is the same as the $mRec_{ins}$ metric with the only difference that the instances of a particular class are only considered. The only class where there is greater variability in the $Rec_{ins}$ metric is the ``other'' class as shown in ~Figure \ref{fig:BFSradiusshape}(h). This is expected since there are instances such as the floor and other shapes that a neighbourhood radius of even $4cm$ will not be sufficient to capture all the points of this class correctly. An ``other'' instance may also correspond to more than one objects grouped together. As such, there will be ``gaps'' between the instances and the instance may not be captured as a whole. This is the reason that for radius $4cm$ the $Rec_{ins}$ metric of ``other'' does not reach 100\%. The same trend is observed for all the \textit{CLOI} facilities. The parameter $\epsilon$ is then confirmed to be $4cm$ based on the above-mentioned observations.

The mean performance metrics (mean precision and recall) for the selected optimal minimum number of points per instance ($\mu$ = 20) given the ground truth class labels for IoU thresholds 25\%, 50\% and 75\% are presented in ~Figure \ref{fig:BFSpreradius} and ~Figure \ref{fig:BFSrecradius} is then presented in Figure~\ref{fig:BFSpreradius} and ~Figure \ref{fig:BFSrecradius}. We tested a range of radii to identify how the distance between separate instances affects the performance of the BFS algorithm. Both mPrec and mRec have an increasing trend with increasing radius ($\epsilon$). The rate of increase of average precision in ~Figure \ref{fig:BFSpreradius} is higher for smaller radii and smaller for higher radii ($\epsilon$ > 4cm). In contrast, the average recall for all IoU thresholds has increasing trend for radii of 1cm and 2cm and then the average recall is decreasing up to 4cm where it converges to 80\% for the warehouse and the petrochemical plant and 60\% for the oil refinery and processing unit. The similarity of the warehouse and the petrochemical plant has been investigated in \cite{agapaki2020cloi} where these facilities have similar normalised point densities. The same trend for the oil refinery and the processing unit is attributed to the similarity in the point density of their point cloud data, which greatly affects the performance and the choice of radius $\epsilon$. The average recall is optimal at a radius of 4cm, therefore we chose this value as the optimal radius for instance segmentation experiments with the BFS algorithm.


We conclude that {\bf the BFS algorithm is more sensitive to changes in the minimum instance size ($\mu$) for $\mu \geq 50$ points rather than changes in the neighbourhood radius ($\epsilon$)} based on the results discussed above. An increase in $\epsilon$ has a relatively linear increase in the performance, both in precision and recall (Figure~\ref{fig:BFSpreradius} and Figure~\ref{fig:BFSrecradius}), whereas an increase in $\mu$ increases precision but reduces recall (Figure~\ref{fig:minInsSize}). This trend was validated by experiments on all the \textit{CLOI} facilities.

\subsection{RESULTS}

The performance of the BFS instance segmentation algorithm was then measured with precision and recall metrics for the optimal parameters of minimum instance size ($\mu$) at 20 points, neighbourhood radius ($\epsilon$) at 4cm and the \textit{class boundary} points to enhance the BFS performance. A class boundary point is defined as a point where its neighbourhood around a radius of 4cm \footnote{The radius selection of 4cm was chosen, since it optimises the instance segmentation performance.} contains at least two points with different \textit{CLOI} class labels. The results for precision and recall per \textit{CLOI} class for the processing unit, the petrochemical plant, the warehouse and the oil refinery are presented in Figure~\ref{fig:BFSNFgtclass}, Figure~\ref{fig:BFSEATONgtclass}, Figure~\ref{fig:BFSTHORgtclass} and Figure~\ref{fig:BFSBPgtclass} respectively. The results for the petrochemical plant indicate that the average precision and recall have very small variation for the same IoU threshold ($\leq 2\%$). {\bf The recall of cylinders for the petrochemical facility is also the highest compared to the other \textit{CLOI} facilities} (61.3\% for IoU=50\% ), which can be attributed to the fact that there were not so many cable trays and twisted conduits that were clustered together or erroneously segmented. This leads to improved instance segmentation of cylinders for the petrochemical facility. {\bf The lowest value of recall of cylinders is for the processing unit} (48.5\% for IoU=50\%). Another trend is evident for all the \textit{CLOI} facilities. The precision performance for most of the \textit{CLOI} classes is higher compared to the recall performance. Particularly, the instance segmentation method when tested on the oil refinery has high precision but reduced recall or in some cases precision and recall have very similar performance (i.e. cylinders or I-beams) as shown in Figure~\ref{fig:BFSBPgtclass}(a) and Figure~\ref{fig:BFSBPgtclass}(b) for $\mu=20$ points. This is attributed to the absence of noisy instances since ground truth class labels are used for instance segmentation in this section. 

The results also show another trend of higher performance for angles, channels, elbows, valves and flanges compared to I-beams and cylinders. The case of I-beams is given as an example. Instance segmentation of I-beams is performed with 59.1\% average precision and 64.2\% average recall when averaging the I-beam instance segmentation results for all \textit{CLOI} facilities. We attribute that to the proximity of I-beam instances, which are frequently located close to or attach each other (i.e. connecting I-beam steel members of a frame). The same trend appears in the case of cylinders. 

We also show the average results of mPrec and mRec for each \textit{CLOI} facility at different IoU thresholds in Figure~\ref{fig:BFSNFgtclass}(c), Figure~\ref{fig:BFSEATONgtclass}(c), Figure~\ref{fig:BFSTHORgtclass}(c) and Figure~\ref{fig:BFSBPgtclass}(c). It is noteworthy that {\bf the greater the IoU threshold, the smaller the value of mPrec or mRec}. This is expected since the lower the threshold is, the easier it is to segment a predicted instance correctly. The performance per \textit{CLOI} class and \textit{CLOI} facility will be discussed in detail in the section that follows.

\subsection{Discussion of instance segmentation results}
\label{instanceDisc}

This section analyses the overall and then the per \textit{CLOI} class instance segmentation performance per facility in isolation. The IoU threshold that is used for the discussion of the results herein is 50\% for consistency with the instance segmentation literature.

Overall, {\bf the instance segmentation method has competitively good performance} (73.2\% mPrec and 71.1\% mRec across the \textit{CLOI} facilities). This is measured in terms of the over-segmented and under-segmented instance point clusters. In other words, the instance segmentation performance is measured in terms of a good match to the ground truth instance point clusters. {\bf The petrochemical plant is the only facility that suffers most from over-segmentation}. The primary reason for that is due to the sparse point density of the TLS dataset. This results in greater over-segmentation of its instances (82.4\% mPrec and 80.9\% mRec, Figure~\ref{fig:BFSNFgtclass}(c)) in comparison to the other \textit{CLOI} datasets. On the other hand, {\bf the oil refinery is the \textit{CLOI} facility that suffers most from under-segmentation}. This is the facility that was densely scanned, however has highly occluded instances. This results in greater under-segmentation compared to the other \textit{CLOI} datasets and greatly affects the instance segmentation performance (65.2\% mPrec and 57.1\% mRec, Figure~\ref{fig:BFSBPgtclass}(c)). 

Figure~\ref{fig:BFS_NF_gt}, Figure~\ref{fig:BFS_THOR_gt}, Figure~\ref{fig:BFS_BP_gt} and Figure~\ref{fig:BFS_EATON_gt} show representative examples of the predicted instance segmented point clusters using the CLOI-Instance segmentation methodology and their comparison to the ground truth segmented instance point clusters. The point clusters are coloured based on the instance point cluster they belong to. In other words, different colour between clusters means these point clusters belong to different instances and there is no direct correlation between the ground truth instance colour and the predicted instance colour. Dashed yellow lines illustrate the instance point clusters to focus on.

{\bf Valves and flanges}. The individual \textit{CLOI} classes for which the CLOI-Instance method has {\bf the highest performance across the \textit{CLOI} datasets} are valves and flanges. The petrochemical plant has the highest precision for valves and flanges (84.3\% and 95.2\% respectively). This is attributed to the majority of flanges being blind flanges with a flat faced surface towards the pipeline they are connected to. The presence of blind flanges easily segments the points around them to those of a cylinder and those belonging to the flange due to the clearly defined boundary points between the flange and the cylinder. The blind flanges in this dataset are used to isolate the piping system or to terminate a pipe as an end. The petrochemical plant dataset has flanges that are not close to each other (they are further than 4cm apart). Those flanges are slip on flanges with flat surfaces and the points of the flange are clearly separable from the points of the pipe. This results in high recall of flanges for this dataset (92\%). The recall of valves is also very high (96\%) for the same dataset. This is attributed to the fact that this type of instances are separated from each other and the likelihood of encountering two valves adjacent to each other in the petrochemical plant dataset is very low. The most frequently encountered types of valves is this dataset are hand-wheel ball valves that are easily distinguishable compared to other types of valves. Similarly, the warehouse has well separated from each other valves resulting in 100\% recall and 68\% precision. The lower precision value can be attributed to some cases where a hand-wheel gate valve and check valve or the sequence of a ball valve and a gate valve are grouped together in the same instance as shown in Figure~\ref{fig:valveflange_gt}(a), Figure~\ref{fig:BFS_THOR_gt}(c) and Figure~\ref{fig:BFS_THOR_gt}(d). Another reason for the slight reduction in the precision of valves in the warehouse dataset is due to over-segmentation of the hand-wheel part of gate valves. This can be attributed to occluded connections between the hand wheel and the body of the valve, or the steam and the hand wheel as shown in Figure~\ref{fig:valveflange_gt}(c).

{\bf The performance of the CLOI-Instance segmentation algorithm on the processing unit and the oil refinery is reduced}. The results show that the processing unit has the same limitations for valve instance segmentation as outlined in the previous facilities. In addition to those, precision is reduced due to highly occluded regions that result in oversegmented instances. The CLOI-Instance performance on flanges is also reduced in the processing unit dataset compared to the other facilities (67.4\% precision and 69.2\% recall), which is attributed to the fact that weld neck flanges are widely prevalent in the processing unit dataset. This means that the boundary between a weld neck flange and a pipe is not a clearly defined flat surface resulting in grouping the two in one instance. A similar case is for the connection between a flange with threaded rods and pipes that are all incorrectly segmented in the same instance as shown in Figure~\ref{fig:valveflange_gt}(e). There are two main pain points for the relatively low performance of the CLOI-Instance methodology on the oil refinery dataset. These are: (a) adjacent valves that are highly occluded and result in undersegmentation and (b) oversegmentation of block and bleed gate valves into their parts (gear actuators and cylinder actuators) as shown in Figure~\ref{fig:valveflange_gt}(g). Figure~\ref{fig:valveflange_gt}(f) shows two characteristic examples of under-segmentation where two pairs of valves are grouped in two instances for the oil refinery dataset.

{\bf I-beams}. The main cause for the {\bf low instance segmentation performance of I-beams} is their direct connectivity with other I-beams in steel frames of the building that clusters all the structural connected members of a frame in one instance. The performance of the method on I-beams is the lowest in comparison to the other structural steel \textit{CLOI} shapes (angles and channels). The average precision and recall of the instance segmentation of I-beams for all the \textit{CLOI} facilities is 59.1\% precision and 64.2\% recall. Figure~\ref{fig:BFS_NF_gt} shows 3D windows with the segmented instances and their respective ground truth instances for the processing unit. Steel I-beams that are usually welded in steel frames cannot be easily segmented by the BFS algorithm and this results in 48.5\% precision and 47.3\% recall. Figure~\ref{fig:BFS_NF_gt}(b), Figure~\ref{fig:BFS_NF_gt}(e) and Figure~\ref{fig:BFS_NF_gt}(g) show that the I-beams are clustered in the same instance, since they are directly connected and they belong to the same class. Instances of I-beams that are welded as parts of steel frames are also grouped together in the oil refinery as shown in Figure~\ref{fig:BFS_BP_gt}(c) and Figure~\ref{fig:BFS_BP_gt}(f). This results in reduced precision and recall of I-beams (50\% and 53.2\% respectively). There are also steel frames composed of I-beams that are grouped together as one instance in the petrochemical plant as shown in Figure~\ref{fig:BFS_EATON_gt}(e). The steel frames in the petrochemical plant dataset are rare to find compared to the other \textit{CLOI} dataset, as such the performance of the CLOI-Instance method on I-beams is higher compared to the other datasets (80\% and 72.4\% respectively).

{\bf Angles and channels}. The instance segmentation performance of angles and channels is {\bf better compared to the instance segmentation performance of I-beams}. The average precision and recall of angles for all the \textit{CLOI} facilities is 81.4\% and 74.3\% respectively. Similarly for channels, the average precision and recall is 77.9\% and 73.1\%. This is attributed to the fact that angles and channels are rare to find attached or welded to each other compared to I-beams. The only exception is the instance segmentation performance of channels in the oil refinery dataset being 73.5\% precision and 56.5\% recall. The primary reason for that is that steel channels in this dataset are connected to each other as parts of stair cases or sidewalks. 

{\bf Cylinders}. The CLOI-Instance method has {\bf relatively low performance on cylinders compared to the other \textit{CLOI} shapes} (56.4\% mean precision and 52.6\% mean recall). This is evident for all the \textit{CLOI} facilities given the following issues of the CLOI-Instance method. 

\begin{enumerate}\setcounter{enumi}{0}
  \item recall of cylinders is affected by complex boundaries that are not clearly defined between weld neck flanges and cylinders ({\bf Cyl1})
  \item reducers, which are classified as ``other'' and their connection to pipes cannot be segmented given that there is no distinctive boundary that separates them  ({\bf Cyl2})
  \item the pipe junctions where cylinders are welded cannot be segmented by the proposed instance segmentation method ({\bf Cyl3})
  \item the conduit that are twisted, bundled in a cable tray or in distances less than 4cm from each other are grouped in the same cylinder instance ({\bf Cyl4}).
  \item electrical boxes that are closely located to each other ({\bf Cyl5})
  \item circular hollow sections of metal stair cases or steel barricades are clustered in one instance ({\bf Cyl 6}).
\end{enumerate}

The {\bf Cyl1} issue is observed in Figure~\ref{fig:BFS_NF_gt}(a) for the processing unit dataset for welded flanges that have points of their connected cylinder misclassified as flange and therefore belong in the same instance. For these cases, the flange and connected cylinder will be inevitably grouped in the same instance. The {\bf Cyl4} issue is prevalent for all the \textit{CLOI} datasets. The processing unit groups different instances together in the case of twisted or bundled conduit that are positioned in cable trays. Similar examples of the {\bf Cyl4} issue are shown for the warehouse dataset, which has conduit that are connected with strut channels and are therefore erroneously predicted as one single instance (Figure~\ref{fig:BFS_THOR_gt}(b) and Figure~\ref{fig:BFS_THOR_gt}(g)). An example of the {\bf Cyl4} issue for the oil refinery dataset in shown in Figure~\ref{fig:BFS_BP_gt}(g). This dataset has the largest number of conduit, which are mostly placed in cable trays and as a result connected. This leads to the lowest precision and recall of cylinders compared to the other three \textit{CLOI} datasets being 39.3\% and 39.4\% respectively. The issue with closely connected conduits is also common in the petrochemical plant dataset as shown in Figure~\ref{fig:BFS_EATON_gt}(a) and Figure~\ref{fig:BFS_EATON_gt}(d). This results in reduced precision and recall of cylinder segmented instances being 64.5\% and 61.3\% respectively. An example of the {\bf Cyl2} issue is observed in Figure~\ref{fig:valveflange_gt}(b) and Figure~\ref{fig:BFS_THOR_gt}(a) where the reducer and pipe are grouped in one cylinder instance for the warehouse dataset. Similarly, a characteristic example of {\bf Cyl3} issue is shown in Figure~\ref{fig:valveflange_gt}(c). An illustrative example in Figure~\ref{fig:BFS_EATON_gt}(i) shows the {\bf Cyl5} issue, where two instances of electric boxes are in contact due to noise.

Over-segmentation is also common among cylinder instances. The warehouse has one large vessel with diameter larger than 1m, however the TLS scanned instance has occlusions that result in segmenting the cylinder to more than one instances (Figure~\ref{fig:BFS_THOR_gt}(f)). This results in relatively low precision of cylinders for the dataset being 61\%. This could be prevented if a denser laser survey was conducted in that area. The oil refinery has similar issues when segmenting cylinder instances as presented in Figure~\ref{fig:BFS_BP_gt}. For example, there are two large vessels with diameter greater than 1m that are split in more than one instances due to its sparsely scanned surface (Figure~\ref{fig:BFS_BP_gt}(f)). 

{\bf Elbows}. The CLOI-Instance method has high performance on elbows across the \textit{CLOI} facility datasets (average precision and recall of 87\% and 71.8\%). There are though three factors that affect the performance, which are:

\begin{enumerate}\setcounter{enumi}{0}
  \item oversegmentation of 180$^{\circ}$ elbows ({\bf El1})
  \item grouping of conduit bends with cylinders, since there is no boundary surface to separate them  ({\bf El2}) and 
  \item clustering of sequences of welded 90$^{\circ}$ elbows in one instance ({\bf El3}).
\end{enumerate}

{\bf The facility that suffers the most from these three limitations is the oil refinery dataset}. Particularly, the recall of method on elbows in this dataset is 56.2\%, the lowest metric compared to the other \textit{CLOI} facilities. An example of the {\bf El3} limitation is shown in Figure~\ref{fig:BFS_BP_gt}(e). The elbows of the processing unit have also slightly reduced recall (60.3\%) compared to the other datasets, which is mostly attributed to the {\bf El1} and {\bf El2} limitation.

On the other hand, there are cases where {\bf the CLOI-Instance segmentation method outperforms the manual instance annotation task}. It is important to note that the class and instance point annotations of the CLOI dataset should not be considered as being perfectly correct. This means that there may inevitably be human annotation errors due to the complexity of these datasets and also the ambiguity of points that are close to boundaries or parts of occluded objects. This is observed especially in the “other” class which is not a CLOI class. Even with the human annotation errors, the performance of the proposed CLOI-instance method is significant and outperforms the human annotation process in specific cases as shown in Figure~\ref{fig:BFS_NF_gt}(h) where a flange and a reducer of the processing unit are correctly predicted. An interesting direction for future research would be to crowdsource the data generation and assess the data quality by validating the task itself (class/instance segmentation) and demonstrate the worker reliability \cite{Liu2019AnAnnotations,Ibrahim2018AutomatedData}. The CLOI-Instance method has superior performance even compared to the manual instance segmentation especially when segmenting the structural frame and the corrugated roof as shown in Figure~\ref{fig:BFS_BP_gt}(d) and Figure~\ref{fig:BFS_BP_gt}(h) for the oil refinery dataset. There are also some cases where the segmentation task using the CLOI-Instance method inevitably leads to over-segmentation. The warehouse has equipment, i.e. pumps that are directly connected to concrete slabs and the floor. These instances are directly connected and therefore all their points are grouped in one instance point cluster (Figure~\ref{fig:BFS_THOR_gt}(e)).

%% file: conclusions.tex
\section{Conclusions}
\label{conclusions}

In the previous section, we evaluated the \textit{CLOI}-Instance segmentation methodology with inputs being the ground truth class segmentation labels. The performance of the CLOI-Instance method depends on two factors: (a) accurate class labels per point and (b) accurate \textit{class boundary} points. The evaluation of the method consisted of the evaluation of the BFS method and the boundary segmentation method using the common metrics of precision and recall. The performance of {\bf the CLOI-Instance method has competitive performance for all the \textit{CLOI} shapes}, achieving 73.2\% mPrec (std=7.8\%) and 71.1\% mRec (std=13.6\%) for all the \textit{CLOI} facilities (Figure~\ref{fig:BFS_BP_gt}(c), Figure~\ref{fig:BFS_THOR_gt}(c), Figure~\ref{fig:BFS_EATON_gt}(c) and Figure~\ref{fig:BFS_NF_gt}(c)), with the state-of-the-art instance segmentation methods achieving 74\% mPrec and 24.9\% mRec as demonstrated in ~Table \ref{table:ASISSGPN}. Thus \textit{CLOI}-Instance achieves significantly higher mRec. The \textit{CLOI}-Instance method achieves very high performance for all \textit{CLOI} facilities especially for classes that are usually not close to each other such as valves or angles. 

The differences in the \textit{CLOI}-Instance performance of \textit{CLOI} facilities are mainly attributed to the point density of the TLS survey \cite{agapaki2020cloi}, which differs dependening on the facility. The \textit{CLOI}-Instance performance per \textit{CLOI} shape is determined by the following two factors:

\begin{enumerate}\setcounter{enumi}{0}
  \item the proximity of an instance point cluster that has the same class label with its closest instance 
  \item the presence of sparsely scanned regions that results in ``gaps'' between points in the instance point clusters.
\end{enumerate}

The \textit{CLOI} classes that are mainly affected due to these factors are cylinders and especially conduit that are located in less than 4cm distance between each other (Figure~\ref{fig:BFSBPgtclass}(a) and ~Figure \ref{fig:BFSBPgtclass}(b)) and I-beams that are welded as parts of steel frames. There are also some special circumstances where the \textit{CLOI}-Instance performance on valves and flanges is affected as presented in Figure~\ref{fig:valveflange_gt} and discussed in the previous section. These are closely located weld neck flanges with pipes or flanges with threaded rods with pipes and over-segmented complex valve shapes such as block and bleed gate or hand-wheel ball valves. The \textit{CLOI}-Instance performance on elbows is also affected in three cases: (a) over-segmentation of 180$^{\circ}$ elbows, (b) grouping of conduit bends with cylinders and (c) sequences of welded 90$^{\circ}$ elbows clustered in one instance. 

However, a direct benefit of the method is that it segments all the points that belong to disjoint instance point clusters and even outperforms the manual instance segmentation task for cases close to the roof of the facility (Figure~\ref{fig:BFS_BP_gt}(b), Figure~\ref{fig:BFS_BP_gt}(d) and Figure~\ref{fig:BFS_BP_gt}(h)) and point clusters close to pipe connections or to steel members/pipe elements (Figure~\ref{fig:BFS_NF_gt}(g), Figure~\ref{fig:BFS_NF_gt}(f) and Figure~\ref{fig:BFS_NF_gt}(h)).

There are some limitations that need to be discussed. The \textit{CLOI}-Instance performance on the oil refinery dataset is reduced (57\% precision and 63\% recall), which is attributed to the high complexity of the dataset, the large number of highly occluded conduits and the large number of connected I-beams. {\bf The boundary segmentation algorithm of the \textit{CLOI}-Instance method has boosted performance on cylinders} by 11\% for the oil refinery dataset, however there are still near-missed instances. This is attributed to the large number of conduits that are placed inside cable trays and also the large number of pipe junctions where the boundaries cannot be clearly defined between instances of the same class (i.e. cylinders). 

We also validated our method using the class segmented clusters of the CLOI-NET method \cite{agapaki2020cloi} as input. It is important to note that noise and outliers were not removed from the input. The mean precision and recall for all the tested CLOI facilities was 48.1\% and 34.7\% respectively. The CLOI shapes that have significantly higher metrics are those with higher class segmentation results. These are cylinders (53.5\% mPrec and 44\% mRec), elbows (66.8\% mPrec) and I-beams (63\% mPrec and 64.3\% mRec). Therefore, it is evident that class segmentation performance heavily impacts instance segmentation performance, so more extensive work improving the class segmentation should be conducted in future research.

\subsection{Contributions}

The contributions of the CLOI-Instance segmentation method are as follows. {\bf Con 2.1} This method discards noisy instances thanks to the boundary segmentation step of the methodology. We observed that the state-of-the-art deep learning instance segmentation networks give many class labels around closely located points in cases of uncertainty for their class label predictions. This influences the instance segmentation branch of the network as well. However, the CLOI-Instance method alleviates this issue by discarding the uncertain class label predictions of the class segmentation network. The method achieves that with the \textit{class boundary} points that enhance the performance of the BFS algorithm. {\bf Con 2.2} This method solves for the first time the instance segmentation problem for the industrial environment settings with reliable performance and achieves remarkable performance (65.2\% mPrec and 57.1\% mRec) when tested with the ground truth class labels as input.

However, the proposed method has its limitations. {\bf Lim 2.1} It might group together instances that are too close to each other. The primary reason for that is that one of the assumptions of the BFS method was that there should be a minimum neighbourhood distance of 4cm between separate distances in order to segment those. Instances that are located too close to each other will be under-segmented. {\bf Lim 2.2} Another limitation of the CLOI-Instance method is that it over-segments instances when there are scanning gaps larger than 4cm within one instance. If an instance is sparsely scanned, this will result in over-segmentation in sub-instances.

Some interesting directions for future research are outlined below:
\begin{itemize}
  \item A graph-cut based method could be used to improve the instance segmentation results instead of the BFS method, especially for instances that are closely located. 
  \item More features, such as point normals or curvatures, could be investigated in the BFS method.
  \item The CLOI-NET \cite{agapaki2020cloi} and instance segmentation proposed in this paper should be validated as a whole framework in order to provide an end-to-end methodology that has as input raw point cloud data and outputs instance point clusters.
  \item 3D object fitting to instance point clusters, which was beyond the scope of this paper can also be investigated. The 3D representation will vary dependent on the use of the gDT and the laser scanned environment.
  \item Another improvement would be to further classify the cylinder point clusters to cylindrical shapes based on their use. For instance, each point could be classified to either pipe, conduit, circular hollow section or vessel.
\end{itemize}

%% file: figures.tex

\begin{figure}[!ht]
\centering
\includegraphics[width=0.95\textwidth]{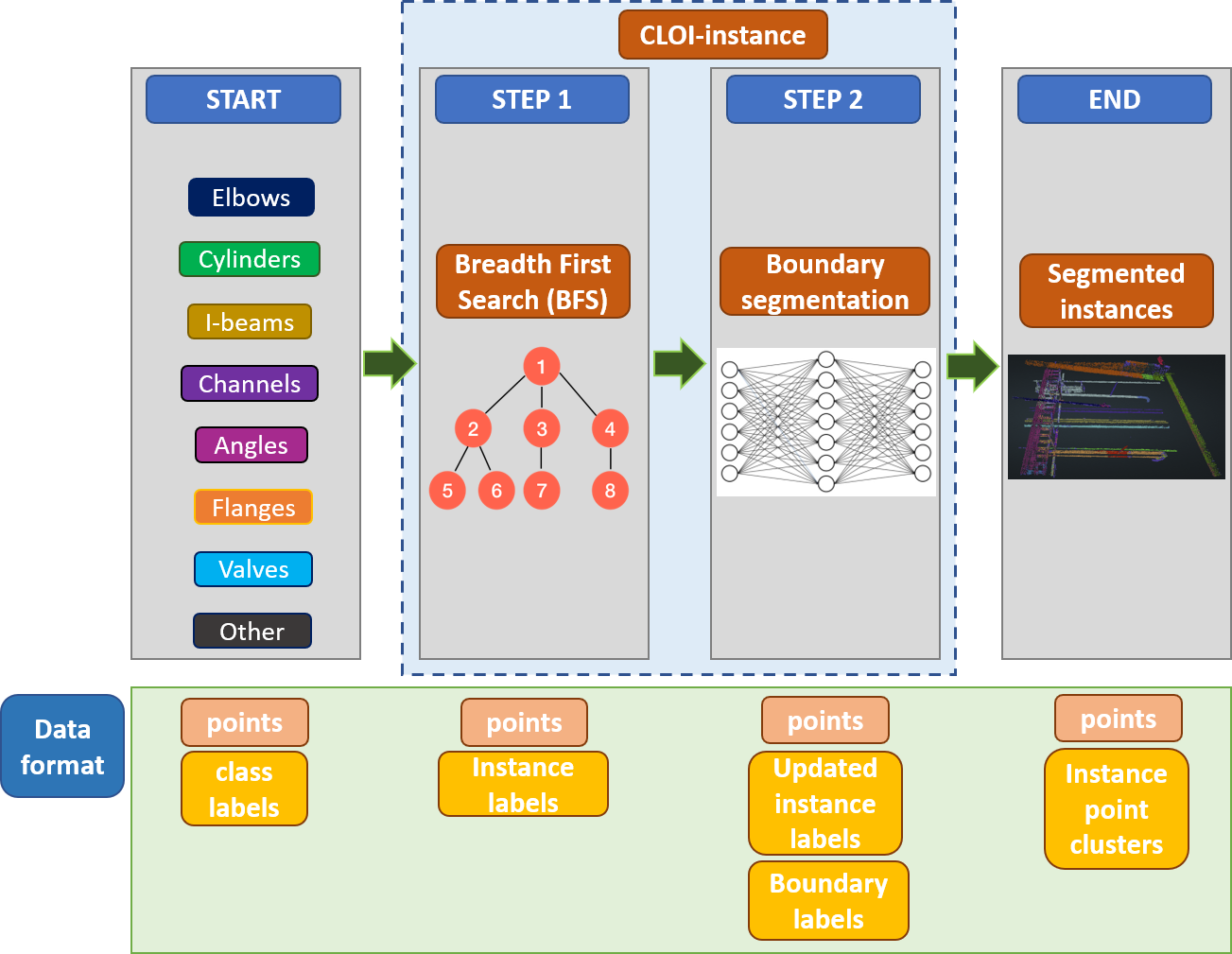}
\caption{Proposed instance segmentation methodology}
\label{fig:insmethodology}
\end{figure}

\begin{figure}[!ht]
\centering
\includegraphics[width=\textwidth]{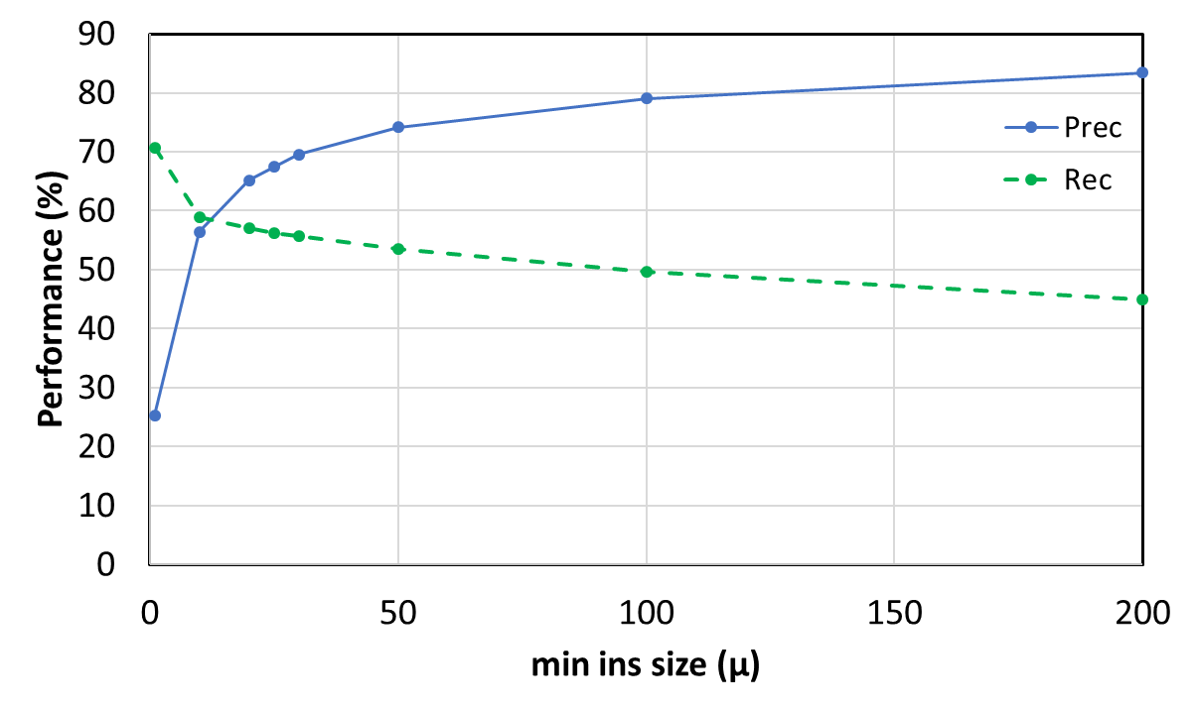}
\caption{Performance of the BFS algorithm with respect to the minimum instance size ($\mu$) for IoU=50\% and $\epsilon=4$ cm. Test on the oil refinery facility.}
\label{fig:minInsSize}
\end{figure}


\begin{figure}[!ht]
\centering
\includegraphics[width=\textwidth]{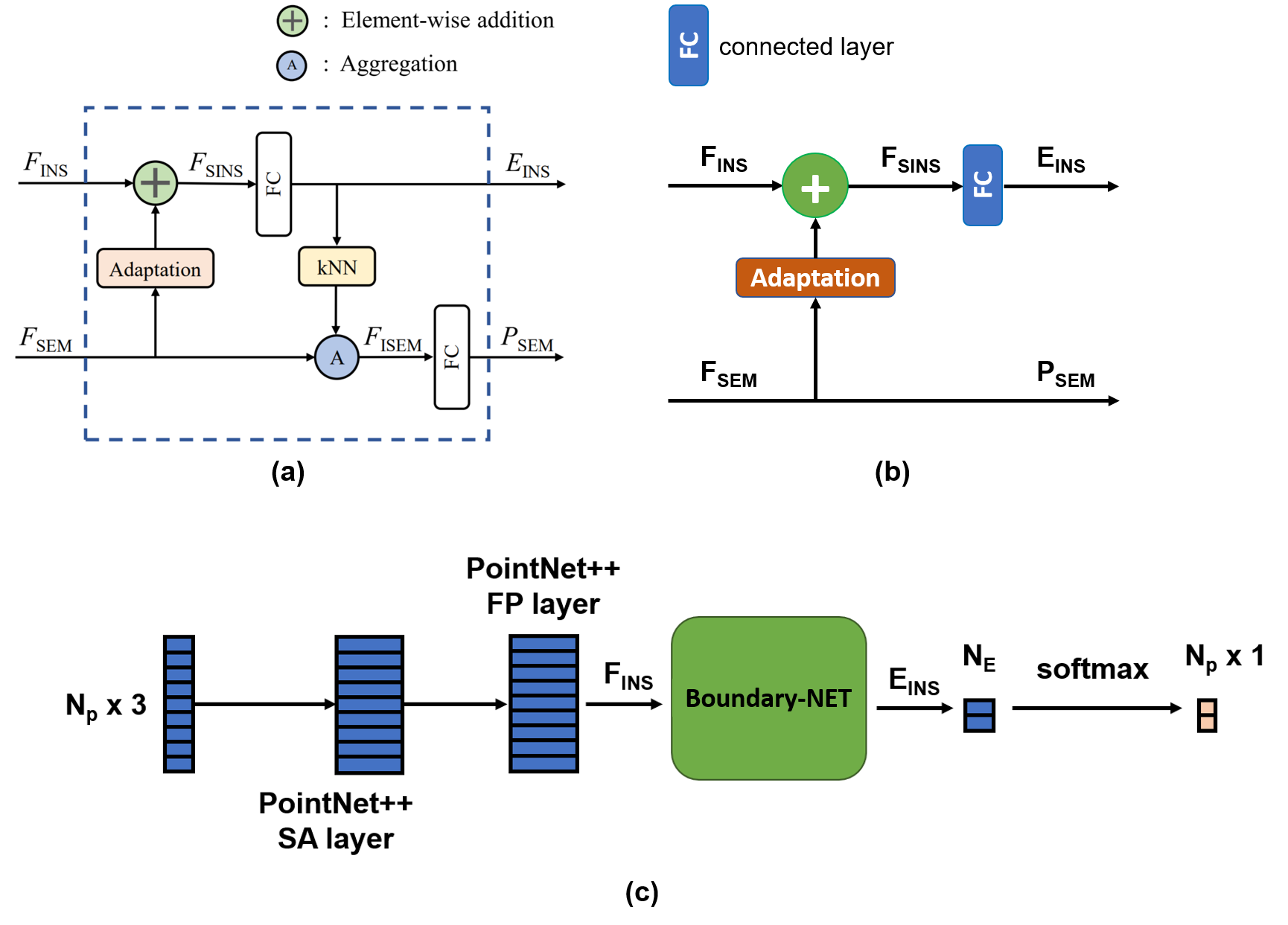}
\caption{Illustration of (a) the original ASIS module and (b) the Boundary-NET module.}
\label{fig:ASIS_network}
\end{figure}


\begin{figure}[!ht]
\centering
\includegraphics[width=\textwidth]{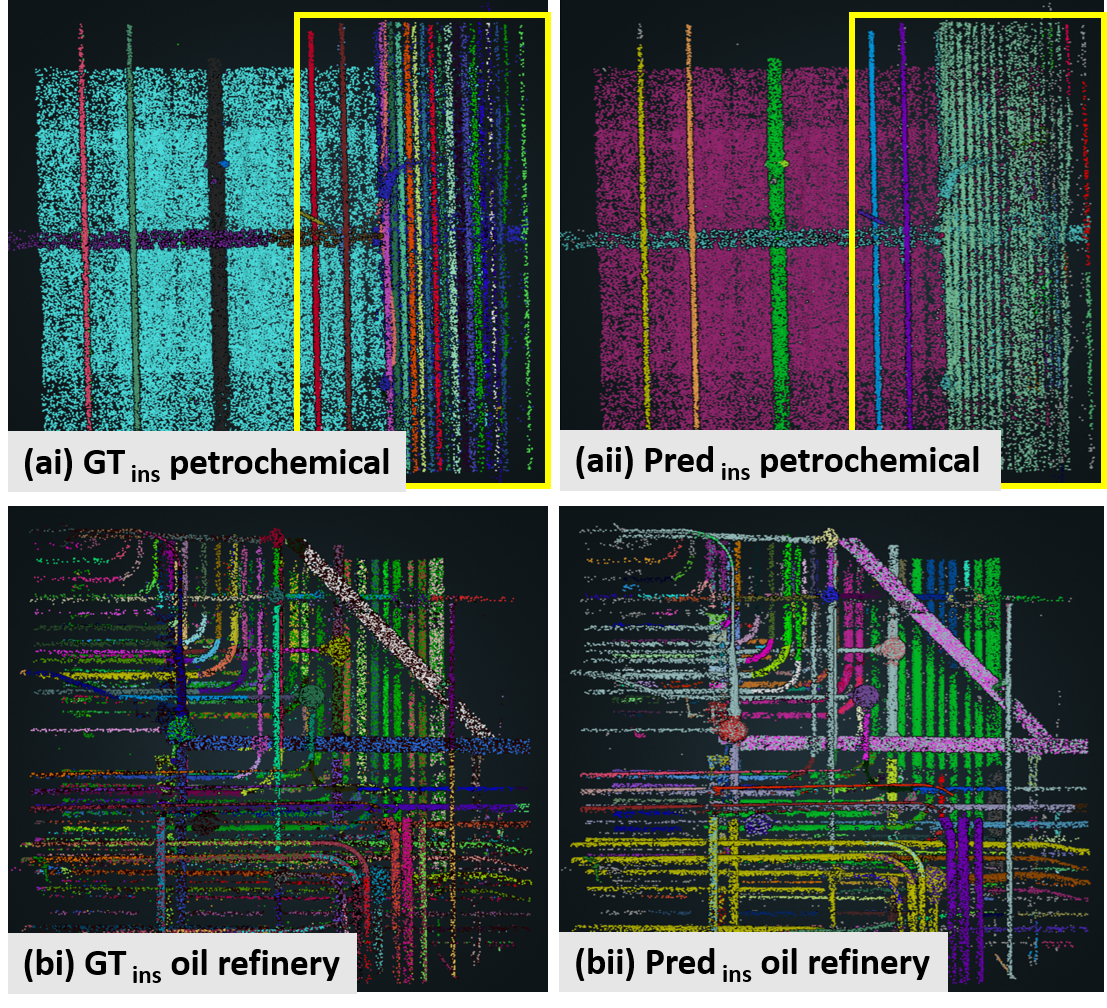}
\caption{Boundary limitations in case of (a) conduit proximity and (b) noisy TLS data. \textcolor{red}{Each color in the figure represents one point cluster instance.}}
\label{fig:BFS_noiseproximity_gt}
\end{figure}



\begin{figure}[t]
\centering
\includegraphics[width=0.85\textwidth]{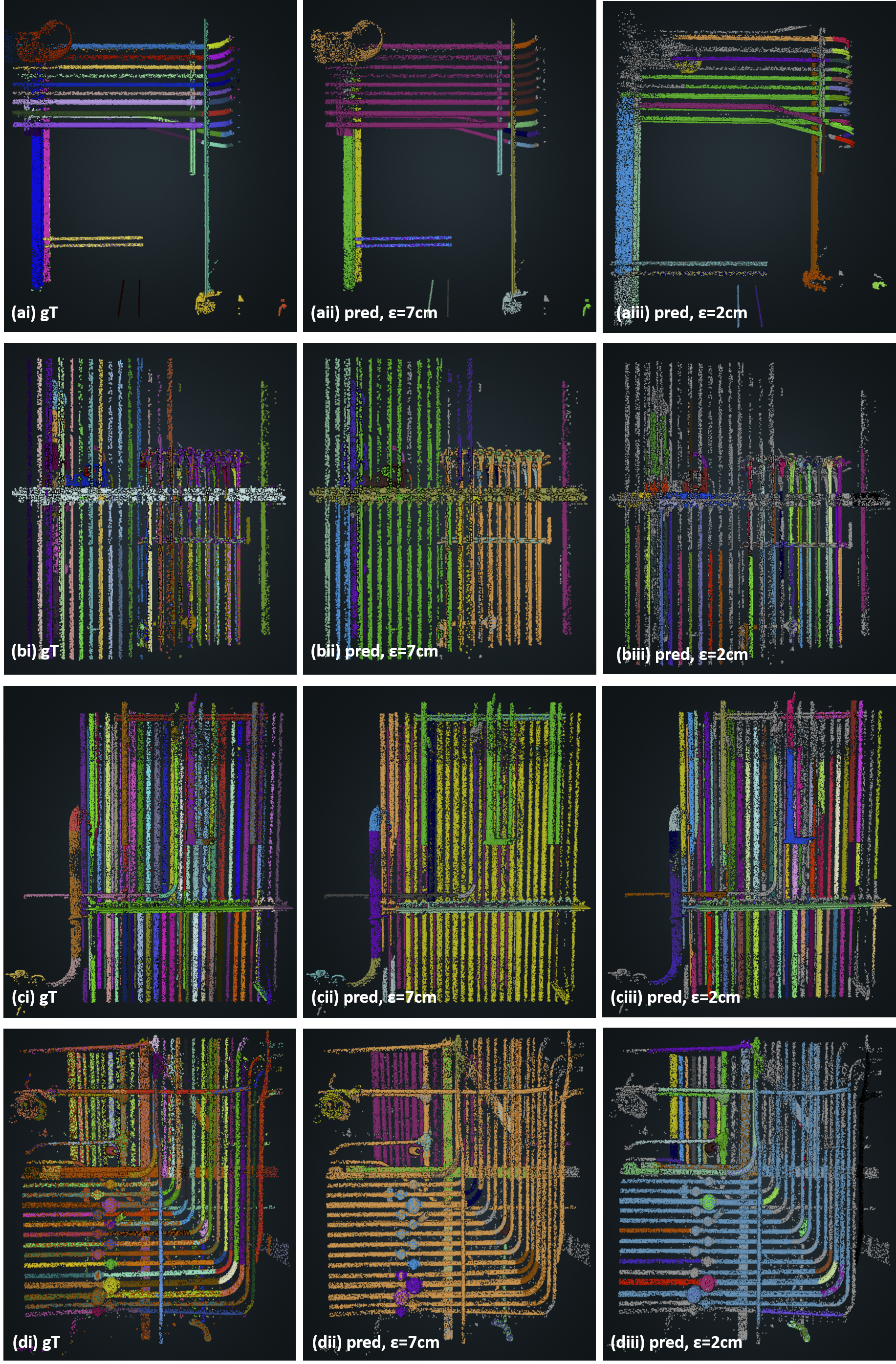}
\caption{(a) Ground truth and predicted instances with BFS for (b) $\epsilon=7cm$ and (c) $\epsilon=2cm$. Results shown for the oil refinery. \textcolor{red}{Each color in the figure represents one point cluster instance. For larger $\epsilon$, we observe under-segmentation of instances, for smaller $\epsilon$ over-segmentation}}
\label{fig:BFS7cm}
\end{figure}

\begin{figure}[t]
\centering
\includegraphics[width=0.7\textwidth]{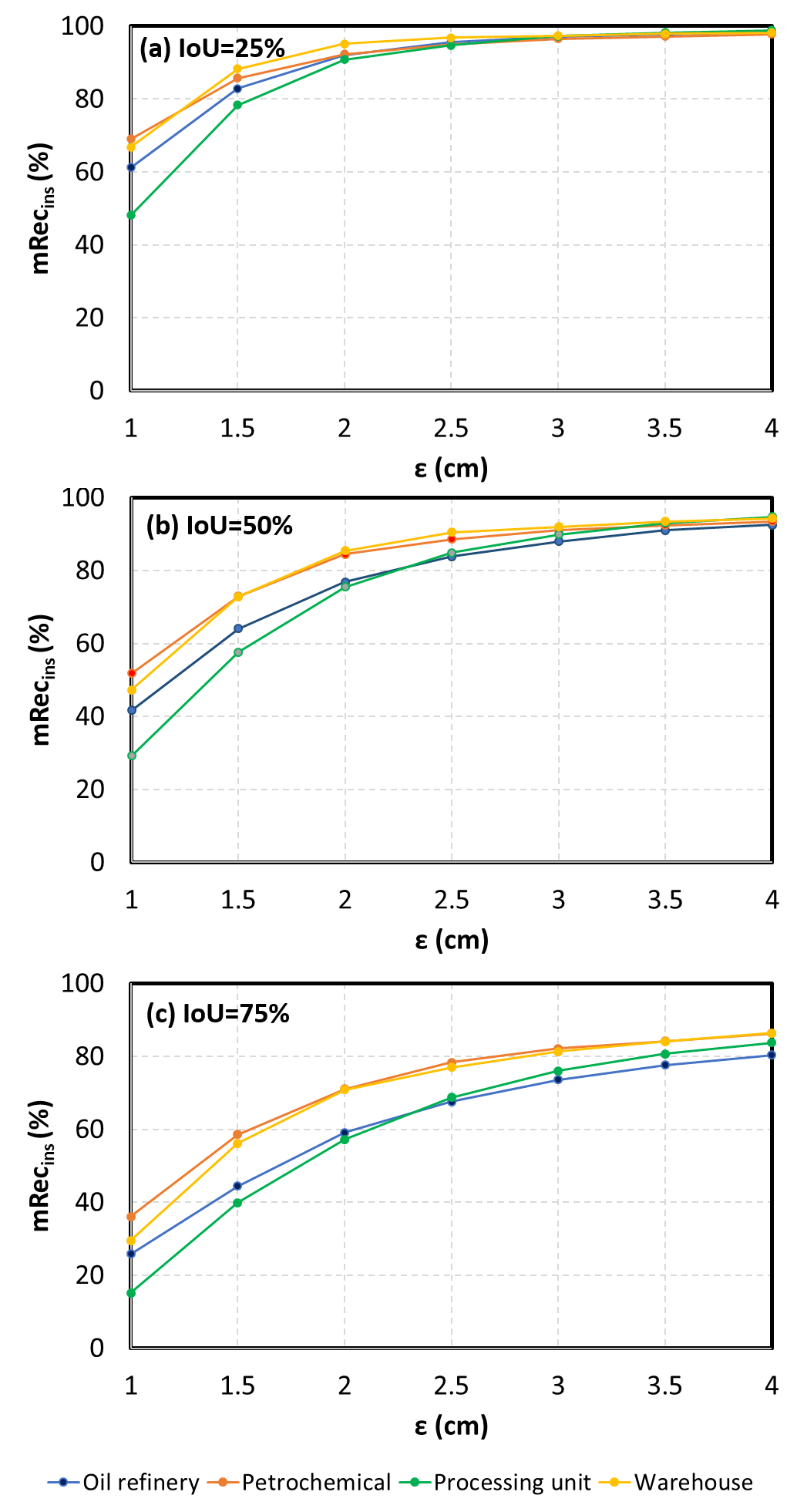}
\caption{BFS radius with respect to $mRec_{ins}$ (\%) for different IoU thresholds for each \textit{CLOI} facility.}
\label{fig:BFSradius}
\end{figure}

\begin{figure}[t]
\centering
\includegraphics[width=1.1\textwidth]{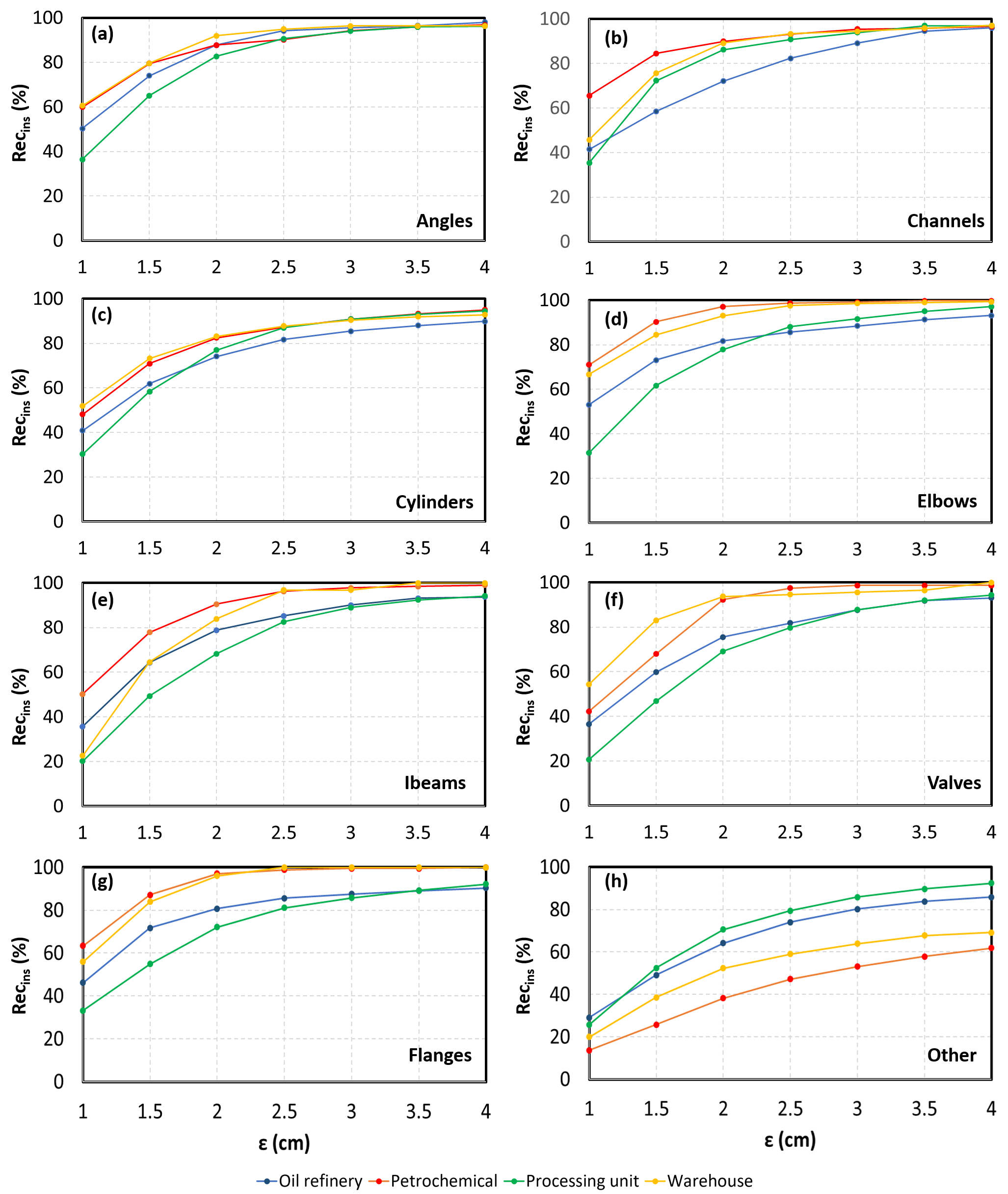}
\caption{BFS radius with Rec (\%) per \textit{CLOI} class for IoU=50\% for each \textit{CLOI} facility.}
\label{fig:BFSradiusshape}
\end{figure}

\begin{figure}[t]
\centering
\includegraphics[width=1.1\textwidth]{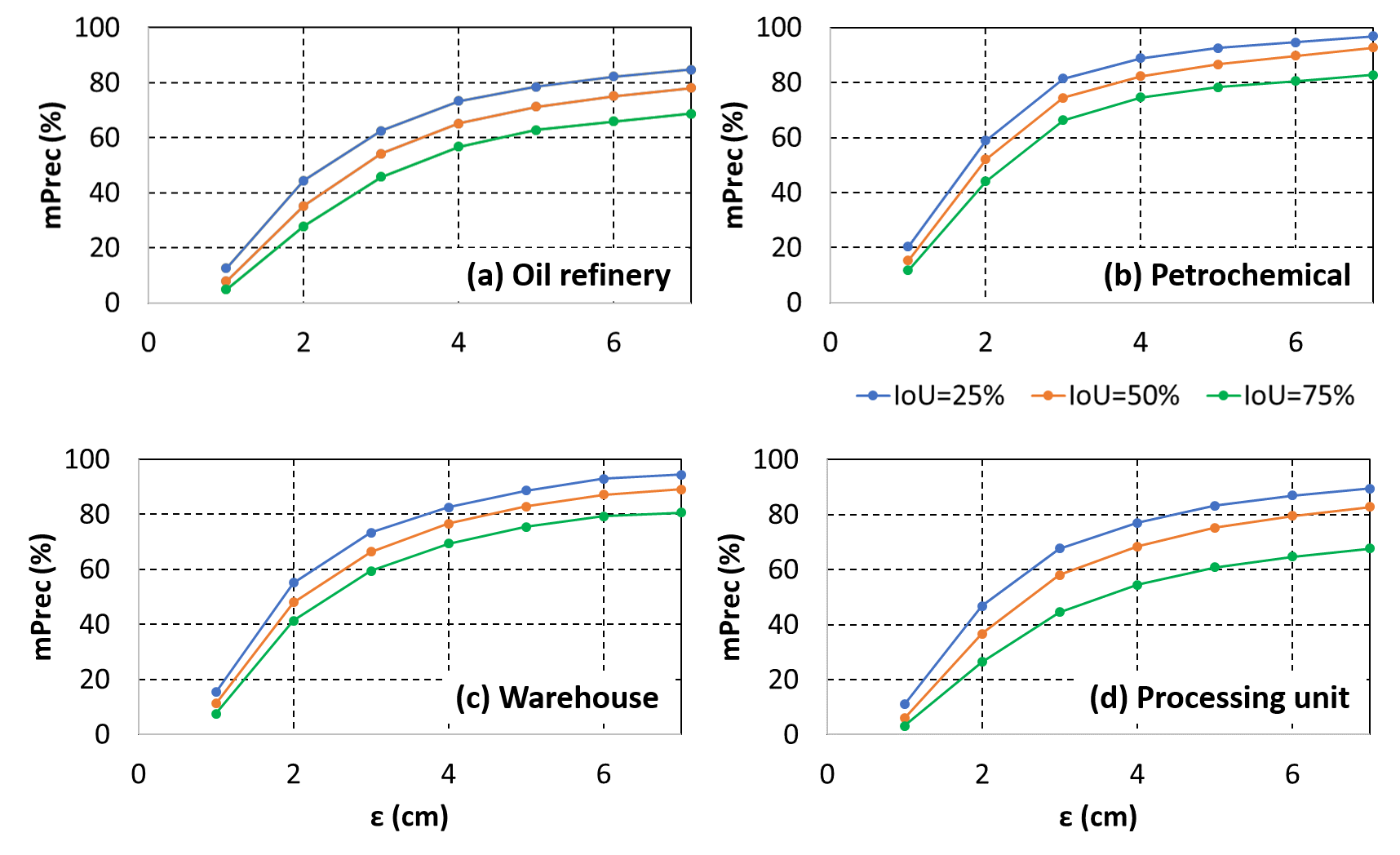}
\caption{Average precision (\%) of the BFS algorithm with respect to the radius ($\epsilon$) for each \textit{CLOI} facility.}
\label{fig:BFSpreradius}
\end{figure}

\begin{figure}[t]
\centering
\includegraphics[width=1.1\textwidth]{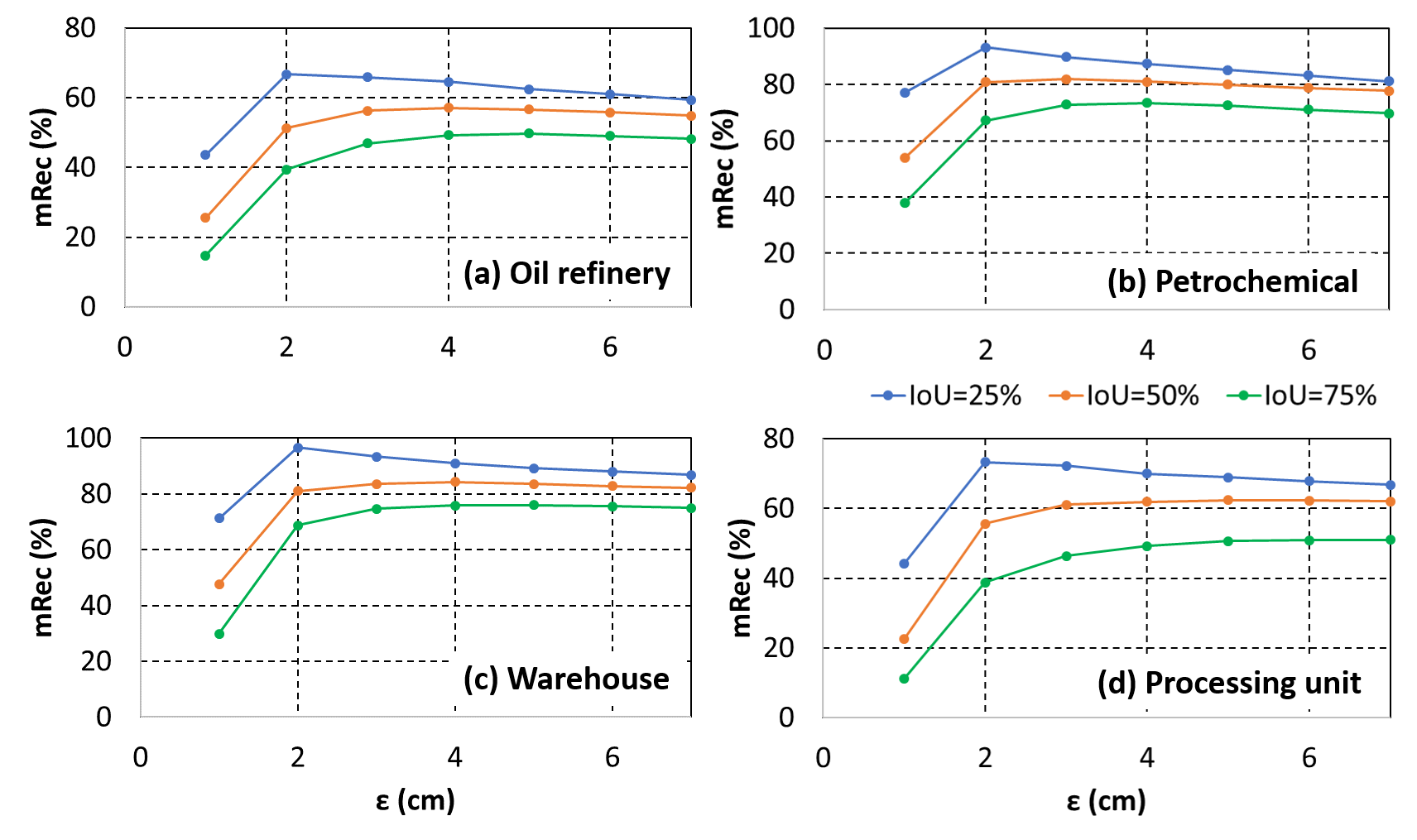}
\caption{Average recall (\%) of the BFS algorithm with respect to the radius ($\epsilon$) for each \textit{CLOI} facility.}
\label{fig:BFSrecradius}
\end{figure}

\begin{landscape}
\begin{figure}[t]
\centering
\includegraphics[width=\textwidth]{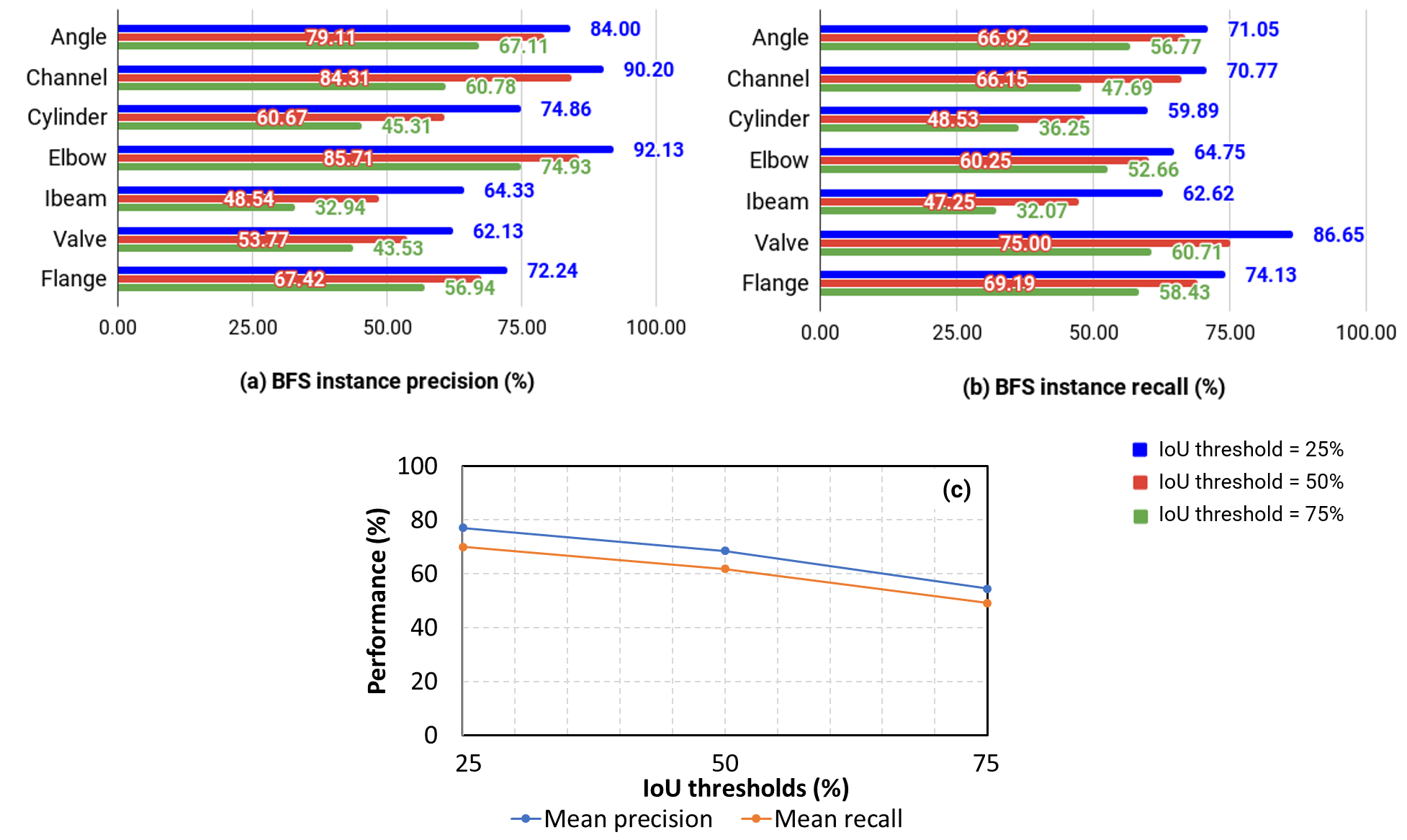}
\caption{(a) BFS instance precision and (b) recall per \textit{CLOI} class and (c) mean precision and recall for different IoU thresholds for the processing unit.}
\label{fig:BFSNFgtclass}
\end{figure}
\end{landscape}

\begin{landscape}
\begin{figure}[t]
\centering
\includegraphics[width=\textwidth]{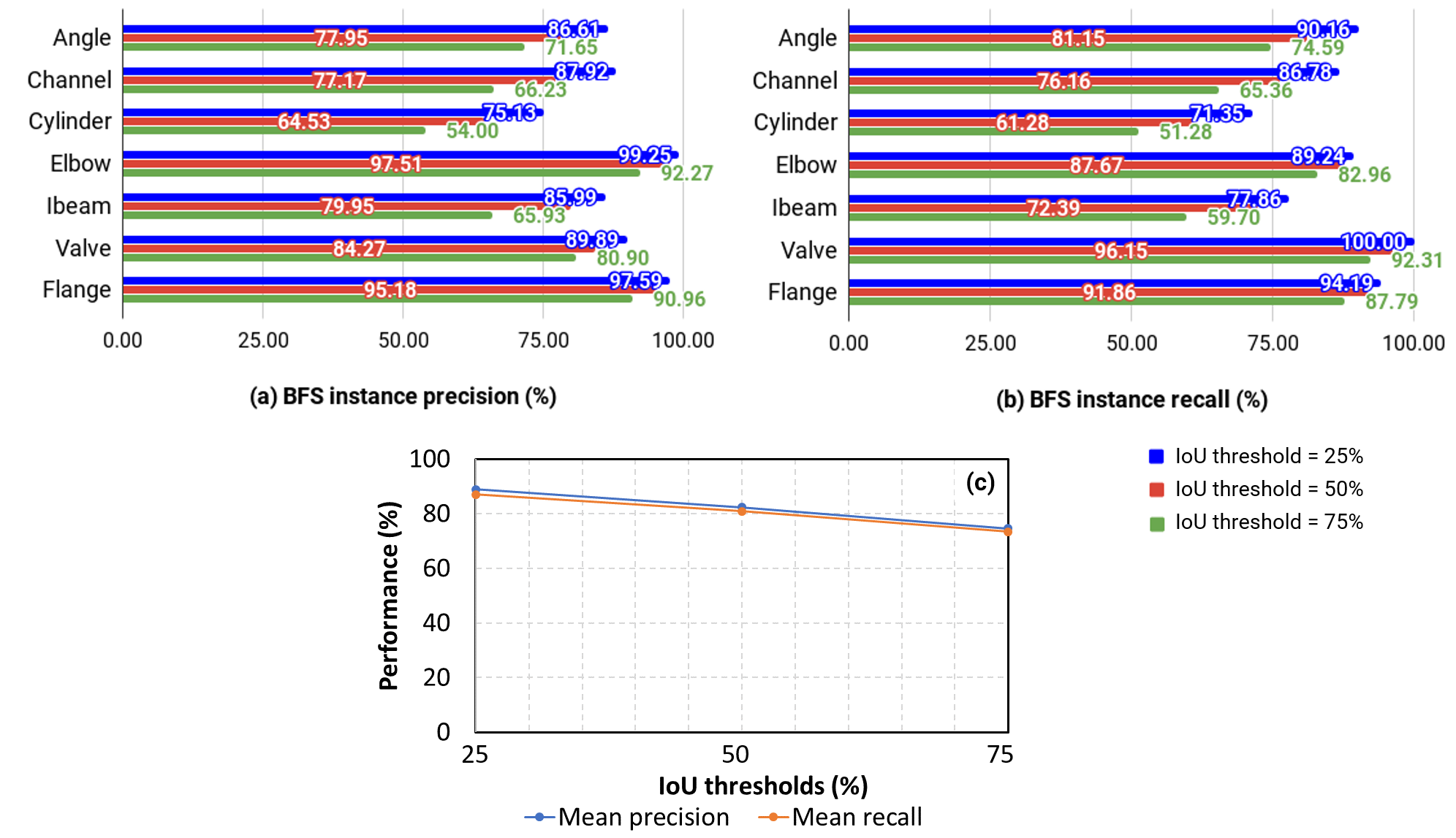}
\caption{(a) BFS instance precision and (b) recall per \textit{CLOI} class and (c) mean precision and recall for different IoU thresholds for the petrochemical plant.}
\label{fig:BFSEATONgtclass}
\end{figure}
\end{landscape}

\begin{landscape}
\begin{figure}[t]
\centering
\includegraphics[width=\textwidth]{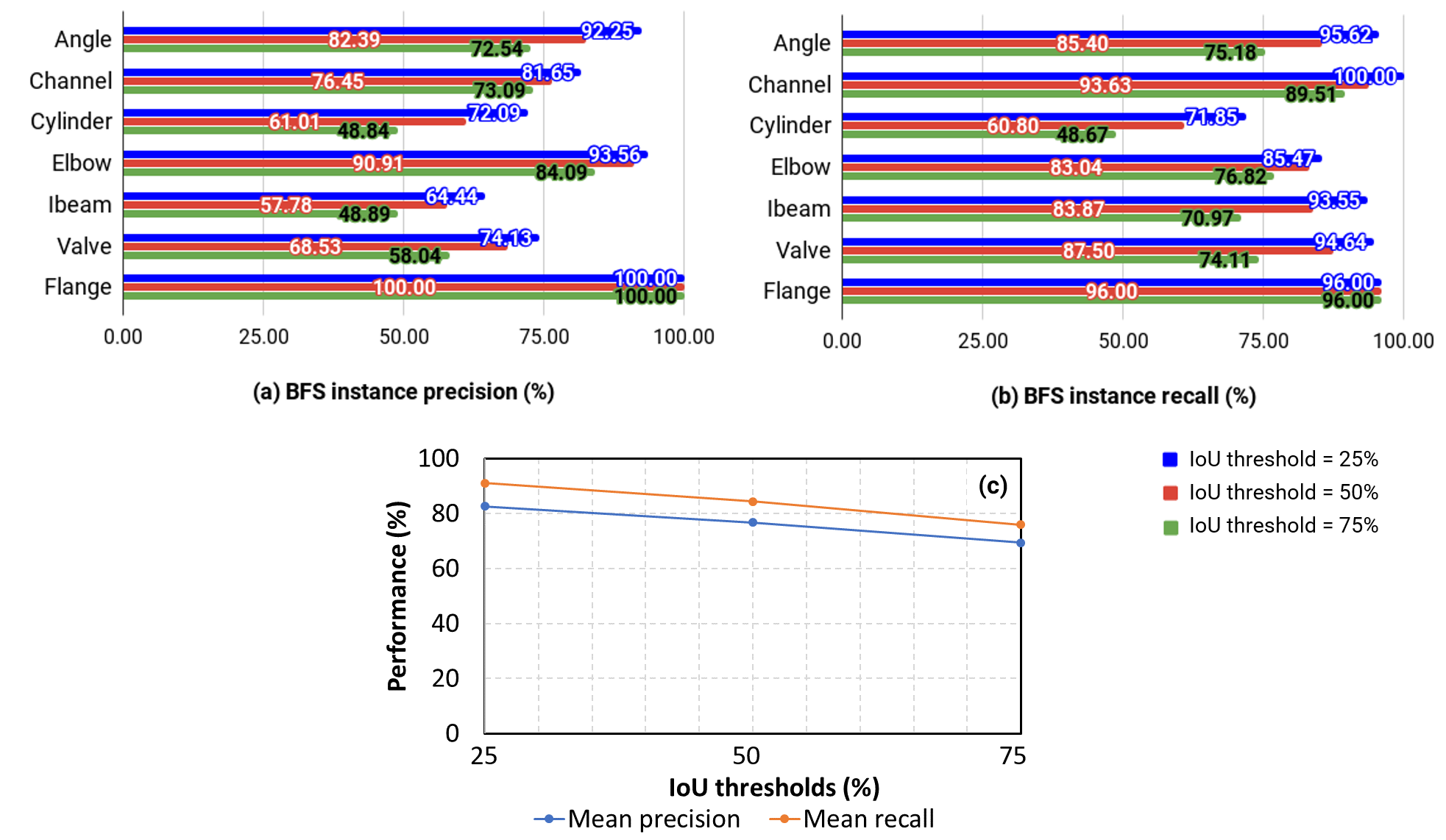}
\caption{(a) BFS instance precision and (b) recall per \textit{CLOI} class and (c) mean precision and recall for different IoU thresholds for the warehouse.}
\label{fig:BFSTHORgtclass}
\end{figure}
\end{landscape}

\begin{landscape}
\begin{figure}[t]
\centering
\includegraphics[width=\textwidth]{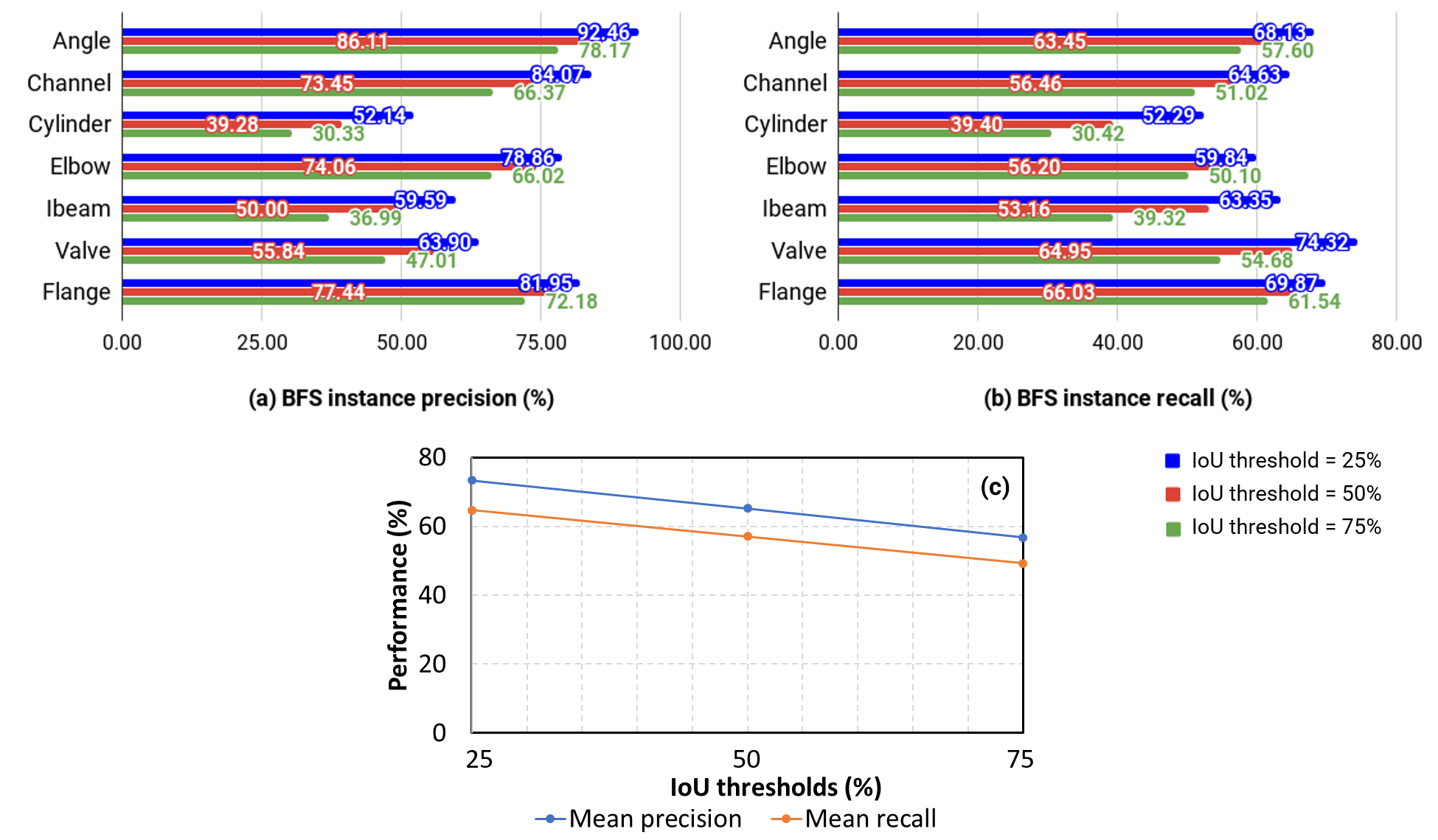}
\caption{(a) BFS instance precision and (b) recall per \textit{CLOI} class and (c) mean precision and recall for different IoU thresholds for the oil refinery.}
\label{fig:BFSBPgtclass}
\end{figure}
\end{landscape}

\begin{figure}[t]
\centering
\includegraphics[width=\textwidth]{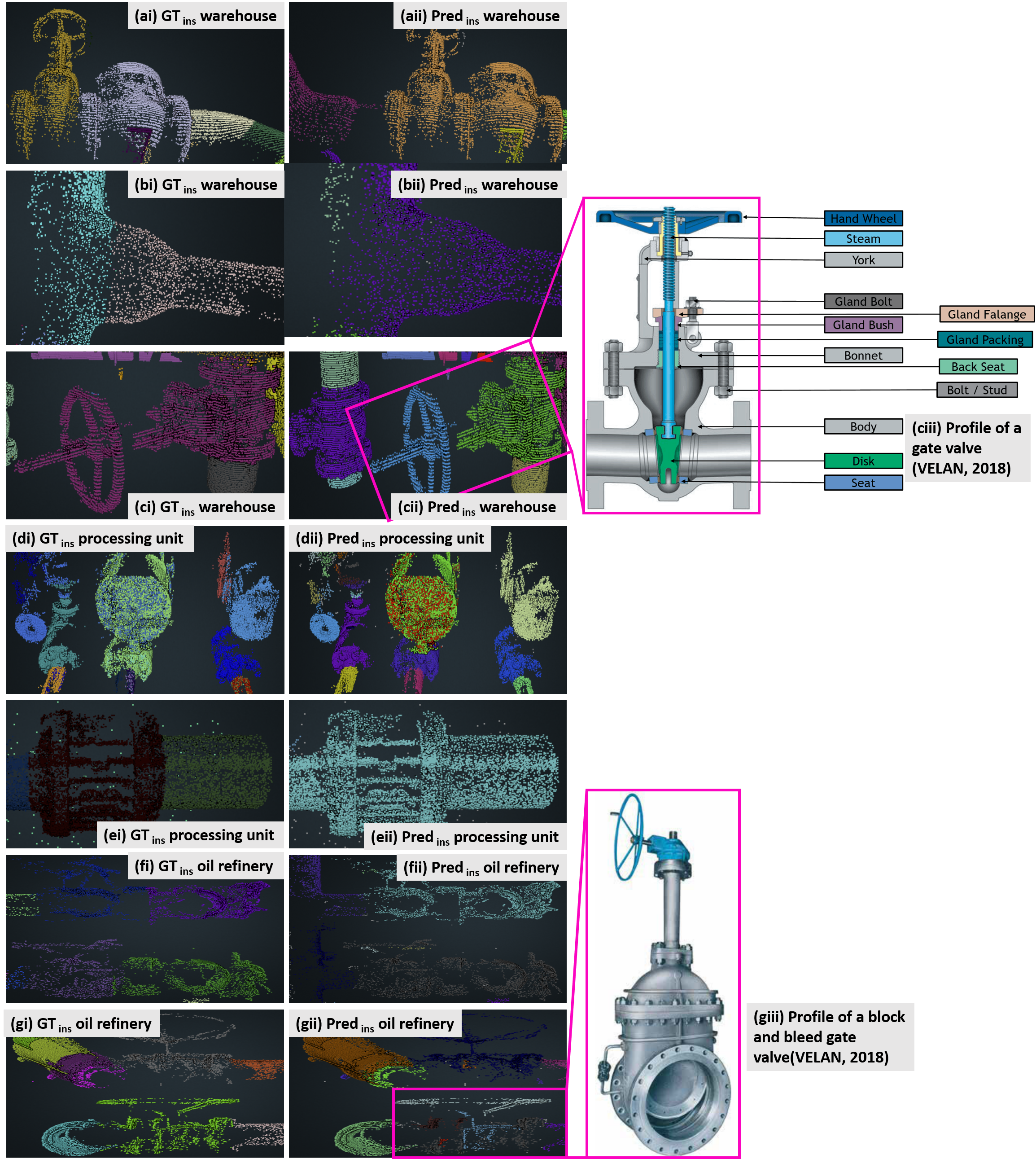}
\caption{(i) Ground truth annotated instances and (ii) predicted instances with BFS alogirthm for representative examples of valves and flanges across the \textit{CLOI} facilities. \textcolor{red}{Each color in the figure represents one point cluster instance.}}
\label{fig:valveflange_gt}
\end{figure}

\begin{figure}[t]
\centering
\includegraphics[width=\textwidth]{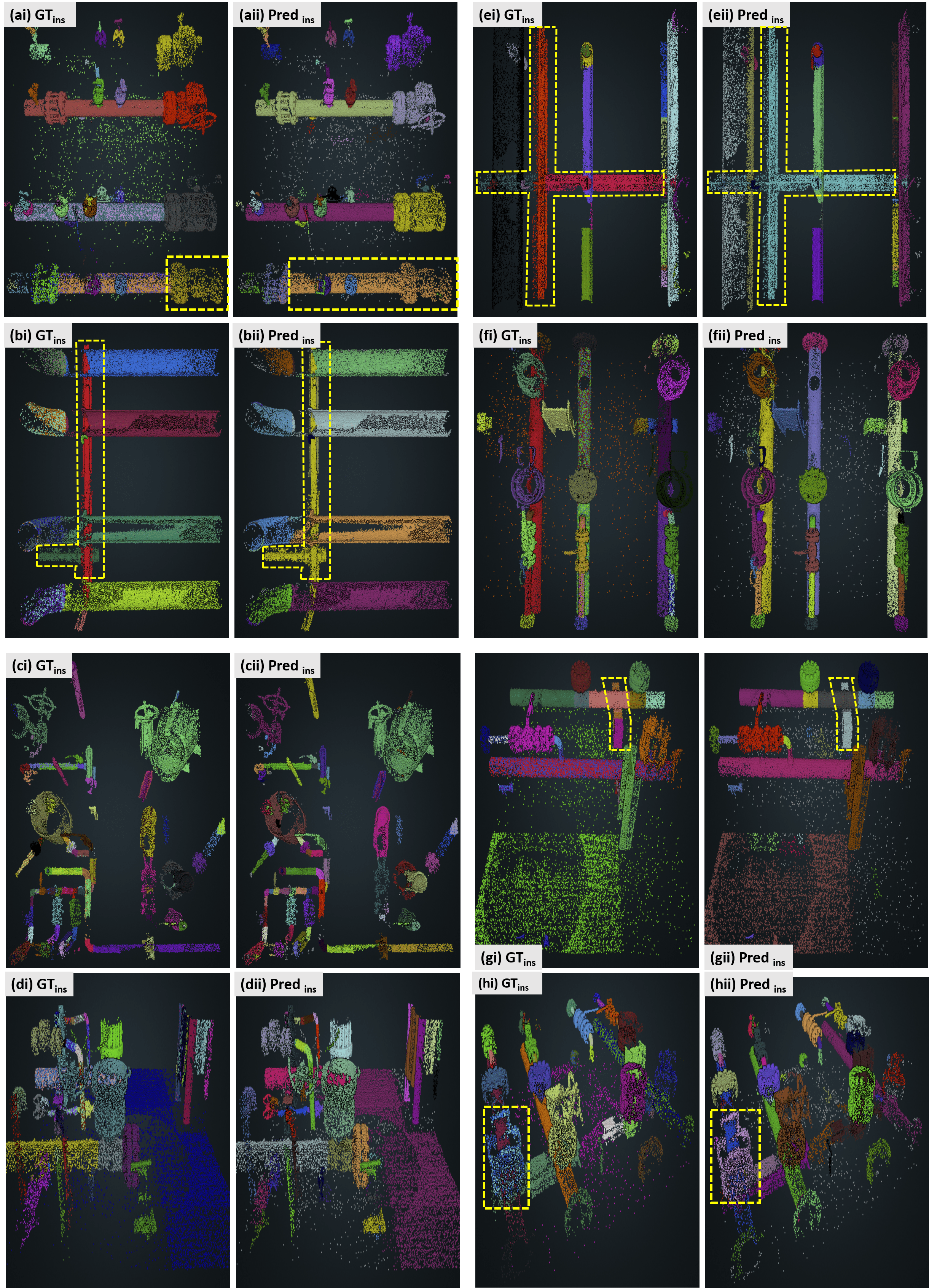}
\caption{(i) Ground truth annotated instances and (ii) predicted instances with BFS algorithm for representative windows from the processing unit.\textcolor{red}{Each color in the figure represents one point cluster instance.}}
\label{fig:BFS_NF_gt}
\end{figure}

\newpage

\begin{figure}[t]
\centering
\includegraphics[width=\textwidth]{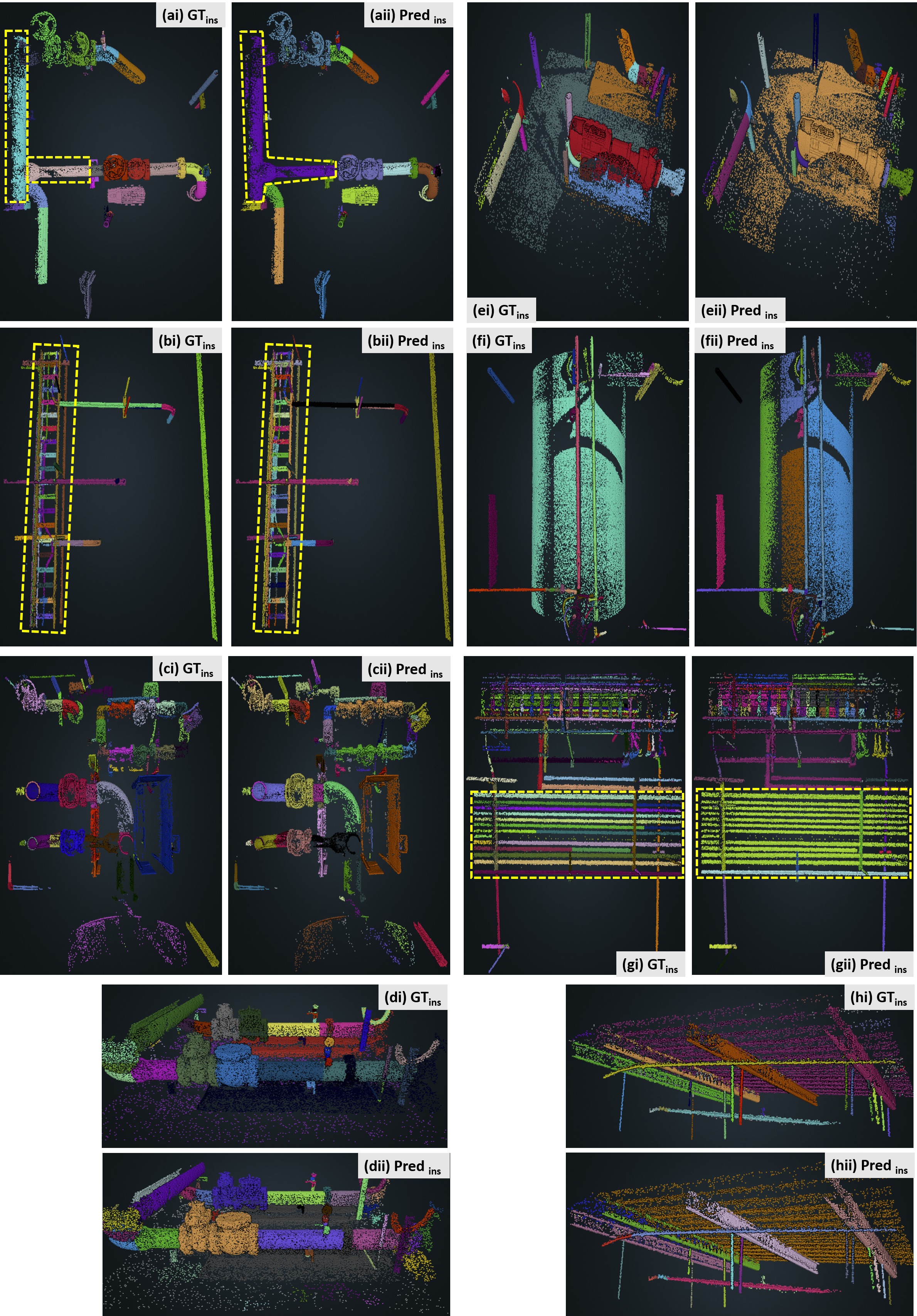}
\caption{(i) Ground truth annotated instances and (ii) predicted instances with BFS alogirthm for representative windows from the warehouse.\textcolor{red}{Each color in the figure represents one point cluster instance.}}
\label{fig:BFS_THOR_gt}
\end{figure}

\begin{figure}[t]
\centering
\includegraphics[width=\textwidth]{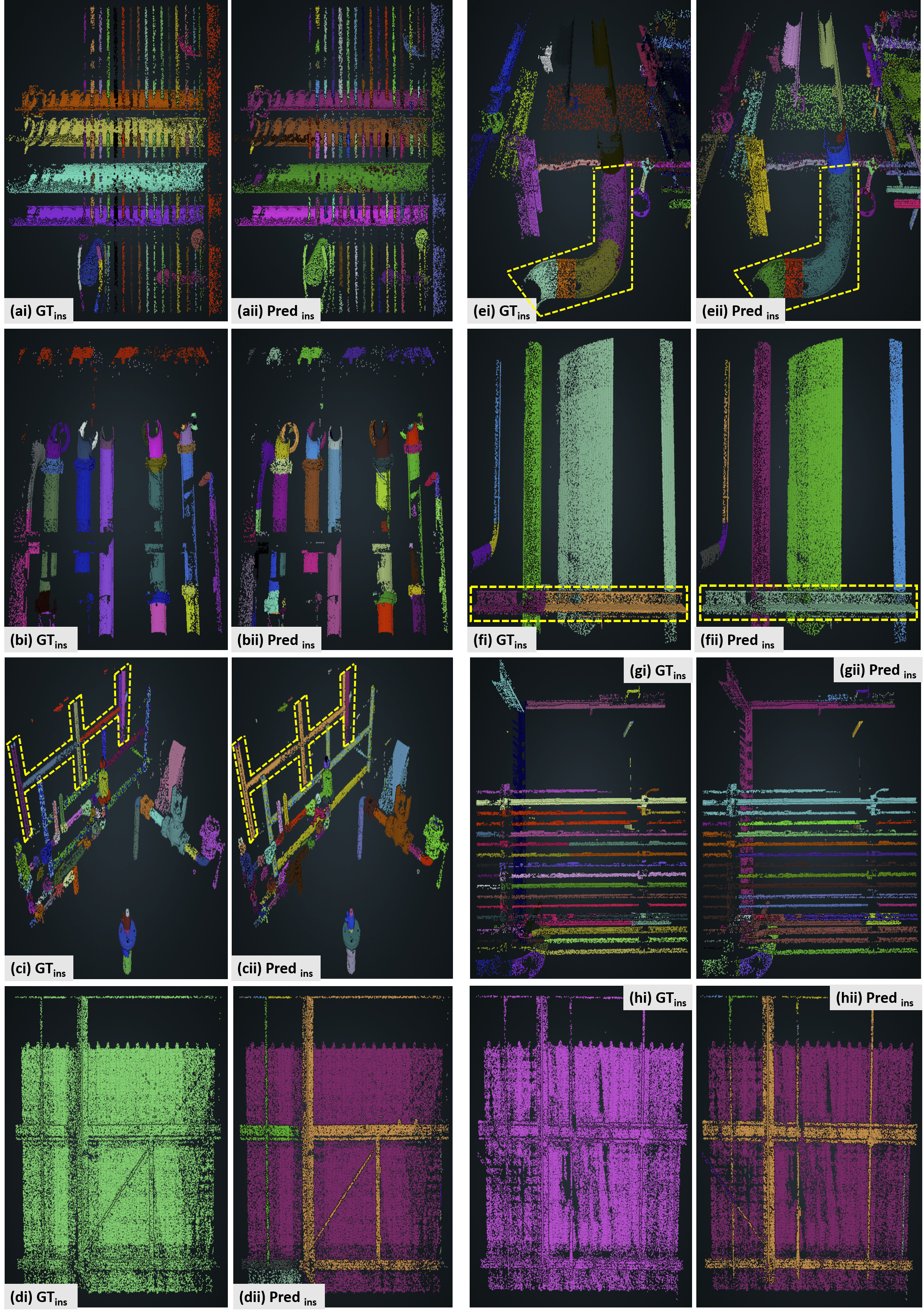}
\caption{(i) Ground truth annotated instances and (ii) predicted instances with BFS alogirthm for representative windows from the oil refinery.\textcolor{red}{Each color in the figure represents one point cluster instance.}}
\label{fig:BFS_BP_gt}
\end{figure}

\begin{figure}[t]
\centering
\includegraphics[width=\textwidth]{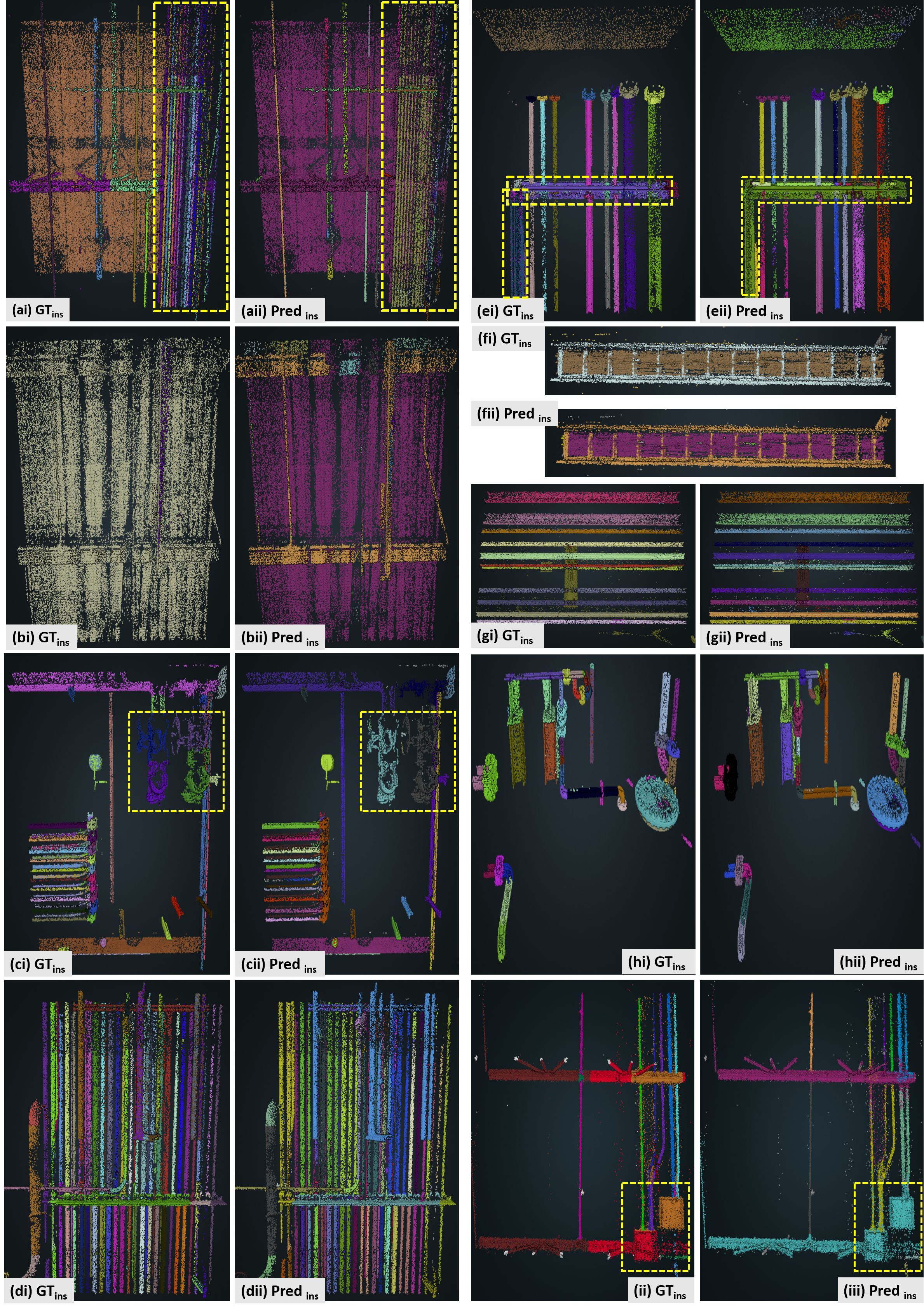}
\caption{(i) Ground truth annotated instances and (ii) predicted instances with BFS alogirthm for representative windows from the petrochemical plant.\textcolor{red}{Each color in the figure represents one point cluster instance.}}
\label{fig:BFS_EATON_gt}
\end{figure}

%% file: Tables.tex


\begin{table}
\centering
\caption{Performance of instance segmentation networks per \textit{CLOI} shape in the oil refinery dataset. Results denoted by (*) use the class segmentation labels of the CLOI-NET method, whereas the rest of the numbers in this table are measured with the ground truth class labels as input.}
\resizebox{\textwidth}{!}{%
    \begin{tabular}{ l l l l l l l l l l }
    \hline\hline
    {\bf Prec (\%)} & {\bf Angles} & {\bf Channels} & {\bf Cylinders} & {\bf Elbows} & {\bf I-beams} & {\bf Valves} & {\bf Flanges} & {\bf Other} & {\bf mPrec (\%)}\\
    \hline
    \ {\bf ASIS \cite{Wang2019AssociativelyClouds}} & 58.7 & 75.9 & 59.7 & 75.6 & 82.8 & 82.9 & 81.9 & 67.4 & 74 (16.7)*\\ 
    \ {\bf SGPN \cite{Wang2018SGPN:Segmentation}} & 34.4 & 48.3 & 23.2 & 14.5 & 11.2 & 15.4 & 4.3 & 79.5 & 22 (5.3)*\\
    \hline
    {\bf Rec (\%) } & {\bf Angles} & {\bf Channels} & {\bf Cylinders} & {\bf Elbows} & {\bf I-beams} & {\bf Valves} & {\bf Flanges} & {\bf Other} & {\bf mRec (\%)}\\
    \hline
    \ {\bf ASIS \cite{Wang2019AssociativelyClouds}} & 18.5 & 27.5 & 11.8 & 17.3 & 43.4 & 27.8 & 27.6 & 87.9 & 24.9 (4.5)*\\ 
    \ {\bf SGPN \cite{Wang2018SGPN:Segmentation}} & 3.5 & 6.9 & 4.6 & 3.1 & 13.2 & 9.6 & 4.3 & 90.7 & 8.4 (6.5)*\\
    \hline\hline
    \end{tabular}
}
\label{table:ASISSGPN}
\end{table}

\begin{table}[!ht]
\centering
\normalsize
\caption{\textit{CLOI} dataset statistics}
    \begin{tabular}{ l l l } 
    \hline\hline
    \ \bf{Warehouse} & \bf{Number of instances} & \bf{Number of points per class}\\
    \hline
    \ Angles & 111 & 157,504\\ 
    \ Cylinders & 910 & 3,171,234\\ 
    \ Channels & 168 & 1,313,851\\
    \ I-beams & 12 & 235,473\\
    \ Elbows & 258 & 256,985\\
    \ Flanges & 21 & 68,128\\
    \ Valves & 85 & 367,642\\
    \ Other & 195 & 4,509,439\\
    \hline\hline
    \ \textit{Oil refinery} & \bf{Number of instances} & \bf{Number of points per class}\\
    \hline\hline
    \ Angles & 211 & 421,223\\ 
    \ Cylinders & 2,347 & 11,776,902\\ 
    \ Channels & 94 & 244,241\\
    \ I-beams & 121 & 2,171,768\\
    \ Elbows & 723 & 912,443\\
    \ Flanges & 215 & 353,881\\
    \ Valves & 202 & 577,367\\
    \ Other & 563 & 10,100,639\\
    \hline\hline
    \ \textit{Petrochemical plant} & \bf{Number of instances} & \bf{Number of points per class}\\
    \hline\hline
    \ Angles & 60 & 591,518\\ 
    \ Cylinders & 1,489 & 10,664,156\\ 
    \ Channels & 264 & 3,060,213\\
    \ I-beams & 140 & 2,841,295\\
    \ Elbows & 376 & 271,947\\
    \ Flanges & 130 & 413,357\\
    \ Valves & 53 & 244,167\\
    \ Other & 828 & 40,240,387\\
    \hline\hline
    \ \textit{Processing Unit} & \bf{Number of instances} & \bf{Number of points per class}\\
    \hline\hline
    \ Angles & 188 & 326,556\\ 
    \ Cylinders & 1100 & 3,812,357\\ 
    \ Channels & 34 & 188,316\\
    \ I-beams & 274 & 2,710,422\\
    \ Elbows & 382 & 392,711\\
    \ Flanges & 229 & 617,760\\
    \ Valves & 341 & 1,213,674\\
    \ Other & 370 & 4,304,156\\
    \hline\hline
    \end{tabular}
\label{table:CLOIstats}
\end{table}

\begin{table}[t]
\centering
\caption{Performance of the BFS algorithm (IoU = 50\%) per \textit{CLOI} shape in the oil refinery dataset ($\mu = 20$ points)}
\resizebox{\textwidth}{!}{%
    \begin{tabular}{ l l l l l l l l l }
    \hline\hline
    {\bf $\epsilon$ = 1cm} & {\bf Angles} & {\bf Channels} & {\bf Cylinders} & {\bf Elbows} & {\bf I-beams} & {\bf Valves} & {\bf Flanges} & {\bf Other}\\
    \hline
    \ {\bf Prec} & 15.5 & 60.8 & 3.7 & 13.6 & 1.9 & 3.9 & 9.9 & 0.5 \\ 
    \ {\bf Rec} & 34.8 & 21.1 & 22.7 & 28.8 & 24.1 & 19.1 & 27.9 & 14.3\\ 
    \hline\hline
    {\bf $\epsilon$ = 2cm} & {\bf Angles} & {\bf Channels} & {\bf Cylinders} & {\bf Elbows} & {\bf I-beams} & {\bf Valves} & {\bf Flanges} & {\bf Other}\\
    \hline
    \ {\bf Prec} & 57.4 & 28.4 & 18.6 & 49.2 & 18.4 & 26.5 & 48.3 & 5.2 \\ 
    \ {\bf Rec} & 58.8 & 46.9 & 40.1 & 50.3 & 50.5 & 51.9 & 60.3 & 41.4 \\  
    \hline\hline
    {\bf $\epsilon$ = 3cm} & {\bf Angles} & {\bf Channels} & {\bf Cylinders} & {\bf Elbows} & {\bf I-beams} & {\bf Valves} & {\bf Flanges} & {\bf Other}\\
    \hline
    \ {\bf Prec} & 76.1 & 53.5 & 31.5 & 66.5 & 37.6 & 44.1 & 69.7 & 14.8 \\ 
    \ {\bf Rec} & 62.3 & 56.5 & 41.6 & 54.4 & 53.6 & 60.1 & 65.7 & 48.7 \\
    \hline\hline
    {\bf $\epsilon$ = 4cm} & {\bf Angles} & {\bf Channels} & {\bf Cylinders} & {\bf Elbows} & {\bf I-beams} & {\bf Valves} & {\bf Flanges} & {\bf Other}\\
    \hline
    {\bf Prec} & {\bf 86.1} & {\bf 73.5} & {\bf 39.3} & {\bf 74.1} & {\bf 50} & {\bf 55.8} & {\bf 77.4} & {\bf 24.4}\\
    {\bf Rec} & {\bf 63.5} & {\bf 56.4} & {\bf 39.4} & {\bf 56.2} & {\bf 53.1} & {\bf 64.9} & {\bf 66.1} & {\bf 49.2}\\ 
    \hline\hline
    {\bf $\epsilon$ = 5cm} & {\bf Angles} & {\bf Channels} & {\bf Cylinders} & {\bf Elbows} & {\bf I-beams} & {\bf Valves} & {\bf Flanges} & {\bf Other}\\
    \hline
    \ {\bf Prec} & 90 & 82.7 & 45.2 & 75.7 & 60.9 & 63.2 & 81.6 & 33 \\ 
    \ {\bf Rec} & 63.1 & 55.1 & 36.8 & 55.3 & 53.9 & 64.9 & 66.9 & 48.7 \\
    \hline\hline
    {\bf $\epsilon$ = 6cm} & {\bf Angles} & {\bf Channels} & {\bf Cylinders} & {\bf Elbows} & {\bf I-beams} & {\bf Valves} & {\bf Flanges} & {\bf Other}\\
    \hline
    \ {\bf Prec} & 91.2 & 86.1 & 48.9 & 77.3 & 68.7 & 67.5 & 86.4 & 39.8 \\ 
    \ {\bf Rec} & 63.7 & 54.4 & 33.6 & 53.7 & 53.9 & 64 & 66.9 & 47.6 \\
    \hline\hline
    {\bf $\epsilon$ = 7cm} & {\bf Angles} & {\bf Channels} & {\bf Cylinders} & {\bf Elbows} & {\bf I-beams} & {\bf Valves} & {\bf Flanges} & {\bf Other}\\
    \hline
    \ {\bf Prec} & 92.3 & 88.8 & 53 & 79.8 & 74.9 & 71.3 & 86.8 & 44.4 \\ 
    \ {\bf Rec} & 63.5 & 53.7 & 31.3 & 52.5 & 54.4 & 63.1 & 65 & 46.4 \\
    \hline\hline
    \end{tabular}
}
\label{table:beforeafterresults1}
\end{table}